\documentclass[journal]{IEEEtran}

\usepackage{array}
\usepackage[caption=false,font=normalsize,labelfont=sf,textfont=sf]{subfig}
\usepackage{textcomp}
\usepackage{stfloats}
\usepackage{url}
\usepackage{verbatim}
\usepackage{graphicx}
\usepackage{cite}

\usepackage{algorithmic}
\usepackage{algorithm}
\usepackage{amsmath,amsfonts}
\usepackage{mathrsfs}
\usepackage{booktabs}
\usepackage{multirow}

\newtheorem{theorem}{Theorem}
\newtheorem{proposition}{Proposition}

\ifCLASSOPTIONcompsoc
  \usepackage[nocompress]{cite}
\else
  \usepackage{cite}
\fi

\ifCLASSINFOpdf
\else
\fi
\begin{document}
\title{Generalizing to Unseen Domains with Wasserstein Distributional Robustness under Limited Source Knowledge} 
\author{Jingge Wang, Liyan Xie, Yao Xie, Shao-Lun Huang and Yang Li 
\IEEEcompsocitemizethanks{
\IEEEcompsocthanksitem Jingge Wang, Shao-Lun Huang and Yang Li are with Tsinghua-Berkeley Shenzhen Institute, Tsinghua University, China.
\IEEEcompsocthanksitem Liyan Xie is with the School of Data Science, The Chinese University of Hong Kong, Shenzhen.
\IEEEcompsocthanksitem Yao Xie is with the School of Industrial and Systems and Engineering, Georgia Institute of Technology.
}
}

\markboth{Journal of \LaTeX\ Class Files,~Vol.~14, No.~8, August~2021}%
{Shell \MakeLowercase{\textit{et al.}}: A Sample Article Using IEEEtran.cls for IEEE Journals}
\IEEEpubid{0000--0000/00\$00.00~\copyright~2021 IEEE}

\maketitle
\begin{abstract}
Domain generalization aims at learning a universal model that performs well on unseen target domains, incorporating knowledge from multiple source domains.
In this research, we consider the scenario where different domain shifts occur among conditional distributions of different classes across domains.
When labeled samples in the source domains are limited, existing approaches are not sufficiently robust. 
To address this problem, 
we propose a novel domain generalization framework called {Wasserstein Distributionally Robust Domain Generalization} (WDRDG),
inspired by the concept of distributionally robust
optimization. We encourage robustness over conditional distributions within class-specific Wasserstein uncertainty sets and optimize the worst-case performance of a classifier over
these uncertainty sets.
We further develop a test-time adaptation module leveraging optimal transport to quantify the relationship between the unseen target domain and source domains to make adaptive inference for target data.
Experiments on the Rotated MNIST, PACS and the VLCS datasets demonstrate that our method could effectively balance the robustness and discriminability in challenging generalization scenarios.
\end{abstract}

\begin{IEEEkeywords}
Domain generalization, distributionally robust optimization, optimal transport, Wasserstein uncertainty set.
\end{IEEEkeywords}




%


\section{Introduction}

%
%
%
%
\IEEEPARstart{I}{n} many practical learning applications, labeled training data are only available from fragmented source domains. It is thus a challenge to learn a robust model for future data that could come from a new domain, with unknown domain shift. One commonly acknowledged solution to this challenge is domain generalization \cite{blanchard2011generalizing}, which aims at learning a model that generalizes well to target domains based on available training data from multiple source domains and in a total absence of prior knowledge about the target domain. A surge of popularity has been seen recently in the application of domain generalization in various fields, such as computer vision \cite{gong2019dlow,shi2020towards,zhou2020learning,zhou2021learning,yue2019domain,volpi2018generalizing,shao2019multi}, natural processing \cite{balaji2018metareg,marzinotto2019robust,stepanov2014towards,fried2019cross}, and reinforcement learning \cite{li2018learning}, etc.

Numerous methods have been developed for learning a generalizable model by exploiting the available data from the source domains,
where the shifts across these source domains are implicitly assumed to be representative of the target shift that we will meet at test time. 
The well-known approaches include learning domain-invariant feature representations through kernel functions \cite{blanchard2011generalizing,muandet2013domain,ghifary2016scatter,blanchard2017domain,grubinger2015domain,li2018domain2,hu2020domain}, or by distribution alignment \cite{zhou2020domain,peng2019moment,motiian2017unified}, or in an adversarial manner \cite{li2018domain,Gong_2019_CVPR,li2018deep,shao2019multi,rahman2020correlation}.
The learned invariance across source domains, however, may not be typical if the unseen target shift is of extreme magnitude.
In this case, forcing distributions to align in a common representation space may result in a biased model that overfits the source domains,
and only performs well for target domains that are similar to certain source domains.

Instead, to explicitly model unseen target domain shifts, meta-learning-based domain generalization methods like MLDG \cite{li2018learning}
divides the source domains into non-overlapping meta-train and meta-test domains, which fails to hedge against the possible target shift beyond the distribution shifts observed in source domains.
Also, these approaches require sufficient source training data to make good meta-optimization within each mini-batch.
Possible domain shift could also been modeled by enhancing the diversity of data 
based on some data augmentations \cite{tobin2017domain}, generating data in an adversarial manner \cite{shankar2018generalizing, volpi2018generalizing, zhou2020deep} or constructing sample interpolation \cite{wang2020heterogeneous,zhou2021mixstyle}.
Learning with limited labeled original samples in this way will weaken their performance, since the new generated data will dominate and the domain shift caused by the artificial data manipulations will largely determine the generalization performance. 

\IEEEpubidadjcol

In this work, we propose a domain generalization framework to explicitly model the unknown target domain shift under limited source knowledge,
by extrapolating beyond the domain shifts among multiple source domains in a probabilistic setting via distributionally robust optimization (DRO) \cite{bagnell2005robust}.
To model the shifts between training and test distributions,
DRO usually assumes the testing data is generated by a perturbed distribution of the underlying data distribution, and the perturbation is bounded explicitly by an uncertainty set. It then optimizes the worst-case performance of a model over the uncertainty set to hedge against the perturbations \cite{sinha2017certifying,gao2018robust,zhu2020distributionally,rahimian2019distributionally}. 
The uncertainty set contains distributions that belong to a non-parametric distribution family, which is typically distributions centered around the empirical training distributions defined via some divergence metrics, e.g., Kullback–Leibler divergence \cite{bagnell2005robust}, or other $f$-divergences
\cite{ben2013robust,duchi2016statistics,namkoong2016stochastic,duchi2021learning}, or Wasserstein distance \cite{sinha2017certifying,blanchet2019robust,lee2018minimax,mohajerin2018data,staib2017distributionally}, etc.
These pre-defined distance constraints of uncertainty sets will confer robustness against a set of perturbations of distributions. 

As a promising tool that connects distribution uncertainty and model robustness, DRO has been incorporated into domain generalization in some works.
Volpi et al. \cite{volpi2018generalizing} augmented the data distribution in an adversarial manner, which appends some new perturbed samples from the fictitious worst-case target distributions at each iteration, and the model is updated on these samples. 
Duchi et al. \cite{duchi2021learning} solves the DRO to learn a model within a $f$-divergence uncertainty set and learns the best radius of the set in a heuristic way by validating on part of the training data.
Let $X$ denote the input feature and $Y$ denote the label.
While the studies by \cite{volpi2018generalizing} and \cite{duchi2021learning} discuss the distributional shifts directly in the joint distribution $P(X, Y)$, 
our work takes a distinct approach by decomposing the joint distribution and establishing class-specific distributional uncertainty sets, which enables us to manage possible varying degrees of distributional perturbations for each class in a more explicit manner.

When labeled training source samples are limited in source domains, the distributional perturbations for each class could vary widely. In such a scenario,
unifying these varying degrees of domain perturbations within a single shared uncertainty set as have been done for the joint distribution is potentially overlooking the inherent differences among classes.
As such, to explicitly examine the distributional shift among classes, we decompose the joint distribution $P(X, Y)=P(X|Y)P(Y)$ and address each part independently. Our primary focus lies in managing the class-conditional shift \cite{zhang2013domain}, under the assumption that there is no shift in the class prior distribution, i.e., the distribution $P(Y)$ stays consistent across all source domains. Furthermore, we also illustrate how our research can be readily expanded to situations that involve a shift in the class prior distribution.
To be more specific, we encode the domain perturbations of each class within a class-specific Wasserstein uncertainty set.
Compared with Kullback–Leibler divergence, Wasserstein distance is well-known for its ability to measure divergence between distributions defined on different probability space, which may happen when the limited samples have no overlap. 
While the classic DRO with one Wasserstein uncertainty set can be formulated into a tractable convex problem \cite{kuhn2019wasserstein},
tractability results for DRO with multiple Wasserstein uncertainty sets for each class are also available \cite{gao2018robust}.

It is crucial to set appropriate uncertainty sets based on training data from multiple source domains for the success of DRO, since they control the conservatism of the optimization problem \cite{mohajerin2018data}. A richer uncertainty set may contain more true target distributions with higher confidence, but comes with more conservative and less practical solution.
More precise uncertainty set incentivizes higher complexity and more difficult solution.
Therefore, uncertainty sets should be large enough to guarantee robustness, but not so large as to overlap with each other. 
We manage to control the discriminability among class-specific uncertainty sets with additional constraints while ensuring the largest possible uncertainty.


When performing classification on data from target domains, 
we conduct a test-time adaptation strategy to further reduce the domain shift and make inference for testing data adaptively.
We employ optimal transport weights to apply the optimal classifier learned from the source distributions on the test sample, which we prove to be equivalent to transporting the target samples to source domains before making the prediction.

In summary, our main contributions include:
\begin{itemize}
    \item We propose a  domain generalization framework that solves the Wasserstein distributionally robust optimization problem to learn a robust model over multiple source domains, where class-conditional domain shifts are formulated in a probabilistic setting within class-specific Wasserstein uncertainty sets. 
    \item To improve upon the original Wasserstein distributionally robust optimization method with heuristic magnitude of uncertainty, we design a constraint that balances robustness and discriminability of uncertainty sets.
    \item 
    We develop a test-time optimal transport-based  adaptation module to make adaptive and robust inferences for samples in the target domain.
    A generalization bound on the target classifier is presented. Experiments on several multi-domain vision datasets show the effectiveness of our proposed framework comparing with the state-of-the-arts.
\end{itemize}

\section{Preliminaries and Problem Setup}
For the common $K$-class classification problem, denote the feature space as $\mathcal{X}\subset \mathbb{R}^d$ and the label space as
{$\mathcal{Y}=\{1,\ldots,K\}$}. 
Let $\phi:\mathcal{X}\rightarrow \Delta_K$ be the prediction function which assigns each feature vector $\boldsymbol{x}$ as class $k$ with likelihood $\phi_k(\boldsymbol{x})$. Here $\Delta_K:=\{\boldsymbol{\xi}\in\mathbb{R}^K:\boldsymbol{\xi}_i\geq0,\sum_{i=1}^K\boldsymbol{\xi}_i=1\}$ denotes the probability simplex. Based on the prediction function $\phi$, the corresponding classifier $\Phi$ maps each feature vector $\boldsymbol{x}$ to the class $\Phi(\boldsymbol{x})=\arg\max_k\left\{\phi_k(\boldsymbol{x})\right\}$ (ties are broken arbitrarily). In the following, we will also use $\phi$ to represent the classifier.

Given training samples $\left\{\left(\boldsymbol{x}_{1}, {y}_{1}\right),\ldots,\left(\boldsymbol{x}_{n}, {y}_{n}\right)\right\}$ drawn i.i.d from the true data-generating distribution over $\mathcal{X}\times\mathcal{Y}$,
we denote the empirical class-conditional distributions for each class as
\[
\widehat{Q}_k:= \frac{1}{|i:y_i=k|}\sum_{i=1}^n\delta_{\boldsymbol{x}_i}1\{y_i=k\}, \ k=1,\ldots,K.
\]
Here, $\delta_{\boldsymbol{x}}$ indicates a Dirac measure centered at $\boldsymbol{x}$ and $1\{\cdot\}$ is the indicator function.
Therefore, $\widehat{Q}_k$ can be viewed as the empirical distribution for training samples within the class $k$. 
In light of \cite{gao2018robust,zhu2020distributionally}, the test distribution of each class is likely to be distributions centered around the empirical class-conditional distribution $\widehat{Q}_k$ within the uncertainty set defined using, for example, the Wasserstein distance. 

The Wasserstein distance \cite{villani2009optimal,peyre2019computational} of order $p$ between any two distributions $P$ and $Q$,
is defined as:
\begin{equation}
    \mathcal{W}_p\left(P, Q\right) = \left(\min_{\gamma\in\Gamma(P,Q)}\mathbb E_{(\boldsymbol{x},\boldsymbol{x}')\sim \gamma}\left[\left\Vert \boldsymbol{x} - \boldsymbol{x}' \right\Vert^p\right]\right)^{1/p},
\end{equation}
where $\Gamma(P,Q)$ is the collection of all joint distributions with the first and second marginals being the distribution $P$ and $Q$, respectively.
We consider the Wasserstein distance of order $p=2$, and the corresponding norm $\left\Vert \cdot \right\Vert$ is set as Euclidean distance.
Thus, we have the test distribution of each class $k$ belongs to the following set:
\begin{equation}
\label{eq:uncertainty_set}
\mathcal{P}_{k}=\left\{P_k\in \mathscr{P}({\mathcal{X}}): \mathcal{W}_2\left(P_k, \widehat{Q}_k\right) \leq \theta_k\right\},
\end{equation}
where $\theta_k\geq 0$ denotes the radius of the uncertainty set and $\mathscr{P}({\mathcal{X}})$ denotes the set of all probability distributions over $\mathcal X$.
A robust classifier $\Phi$ (or equivalently the prediction function $\phi$) can be obtained by solving the following minimax optimization problem:
\begin{equation}
\min _{\phi:\mathcal{X}\rightarrow \Delta_K} \max _{P_{k} \in \mathcal{P}_{k}, 1\leq k \leq K }\Psi\left(\phi; P_1,\ldots,P_{K}\right),
\label{eq:robustHT}
\end{equation}
where $\Psi\left(\phi; P_{1},\ldots, P_{K}\right)$ is the total risk of the classifier $\phi$ on certain distributions $P_1,\ldots,P_{K}$. 
The inner maximum problem refers to the worst-case risk over uncertainty sets $\mathcal{P}_1,\ldots,\mathcal{P}_{K}$.
Suppose $(\phi^*;P_1^*,\ldots,P_{K}^*)$ is an optimal solution pair to the saddle-point problem \eqref{eq:robustHT}, then $P_1^*,\ldots,P_{K}^*$ are called the least favorable distributions (LFDs) \cite{huber1965robust}, 
and $\phi^*$ induces the optimal classifier that minimizes the worst-case risk.

The likelihood that a sample is misclassified is usually taken as the risk, i.e., $1-\phi_k(\boldsymbol{x})$ for any sample $\boldsymbol{x}$ with real label $k$. Specially, when assuming the simple case with equal class prior distributions $\mathbb{P}(y=k)=1/K,k=1,\ldots,K$ for all classes, the total risk of misclassifying data from all $K$ classes is 
\begin{equation}
\Psi\left(\phi; P_{1}, \ldots P_{K}\right)=\sum_{k=1}^{K} \underset{\boldsymbol{x} \sim P_{k}}{\mathbb{E}}\left[1-\phi_{k}(\boldsymbol{x})\right].
\label{eq:risk_function}
\end{equation}
However, in a more general classification problem, to compensate for the possible class imbalance scenario, a series of class-weighting methods assign different weights to misclassifying samples from different classes \cite{zhou2005training,scott2012calibrated}.
One of the most natural approaches is to incorporate the class prior distributions $\mathbb{P}(y=k)$ of each class into the risk function \cite{aurelio2019learning,xu2020class} as
\begin{equation}
\Psi\left(\phi; P_{1}, \ldots P_{K}\right)=\sum_{k=1}^{K} \mathbb{P}(y=k) \underset{\boldsymbol{x} \sim P_{k}}{\mathbb{E}}\left[1-\phi_{k}(\boldsymbol{x})\right],
\label{eq:weighted_risk_function}
\end{equation}
which is a general form of \eqref{eq:risk_function}.

In domain generalization problems, we have access to $R$ source domains $\{\mathcal{D}^{s_r}\}_{r=1}^R$, with training samples $\left\{\left(\boldsymbol{x}_{1}, \boldsymbol{y}_{1}\right),\ldots,\left(\boldsymbol{x}_{n_r}, \boldsymbol{y}_{n_r}\right)\right\}$ from the $r$-th source domain drawn i.i.d from the joint distribution ${P}^{s_r}$ on $\mathcal{X}\times\mathcal{Y}$.
The goal is to learn a robust classifier that performs well on the unseen target domain $\mathcal{D}^t$,
which contains instances from the joint distribution ${P}^t$. 
For each class $k$, denote the empirical class-conditional distributions in source domain $\mathcal{D}^{s_r}$ and target domain $\mathcal{D}^{t}$ as $\widehat{Q}^{s_r}_k$ and $\widehat{Q}^{t}_k$, respectively.
Instead of constructing uncertainty sets relative to the  empirical (training) distributions of a single domain as in the classic DRO formulation, 
we need to set the uncertainty sets using distributions $\widehat{Q}_k^{s_r}$ from multiple source domains, which is detailed in the next section.

\begin{figure*}[!t]
\centering
\subfloat[Construction of uncertainty sets.]{\includegraphics[width=0.45\linewidth]{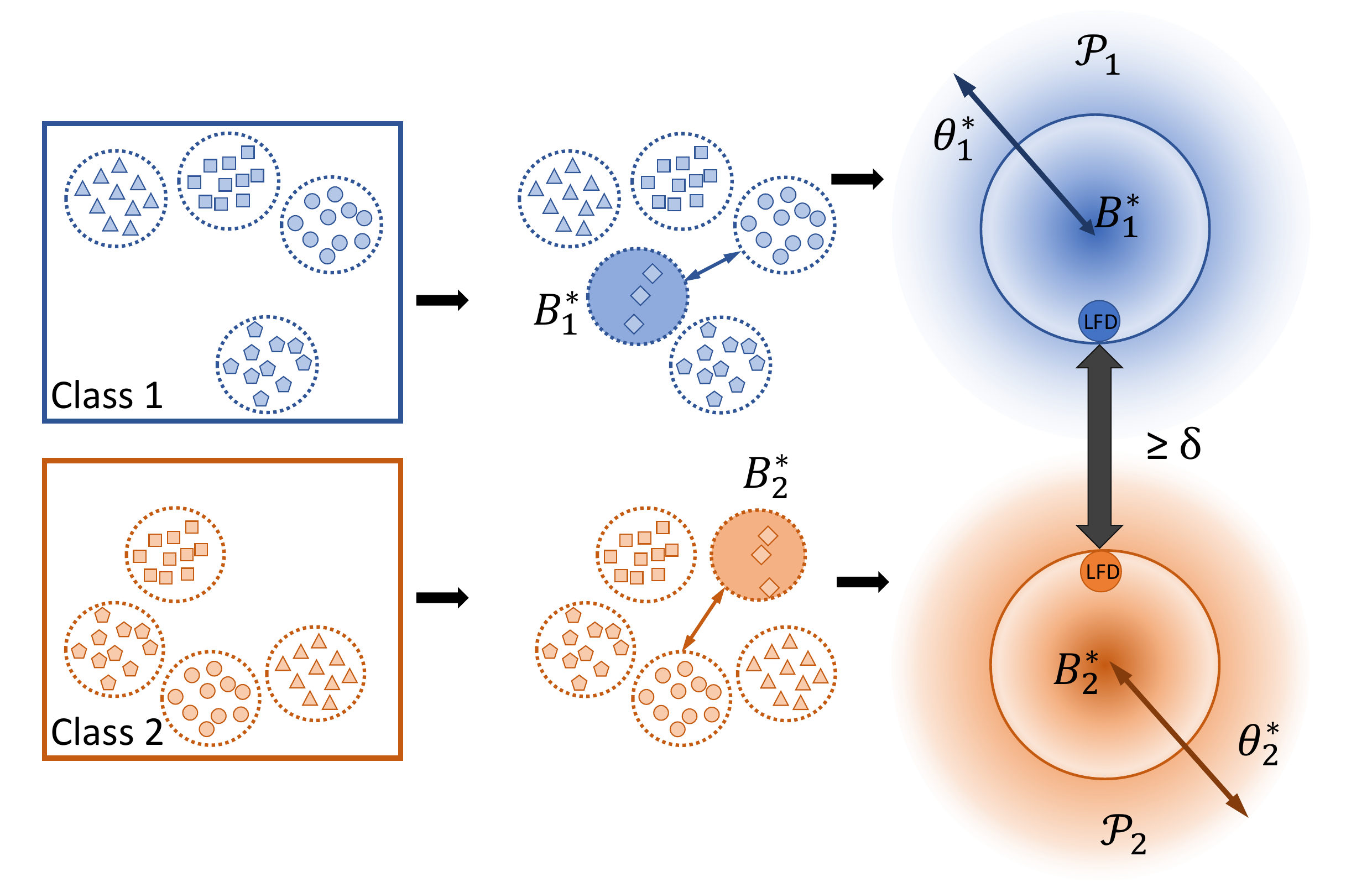}
\label{subfig:a}}
\hfil
\subfloat[Distributionally robust optimization.]{\includegraphics[width=0.45\linewidth]{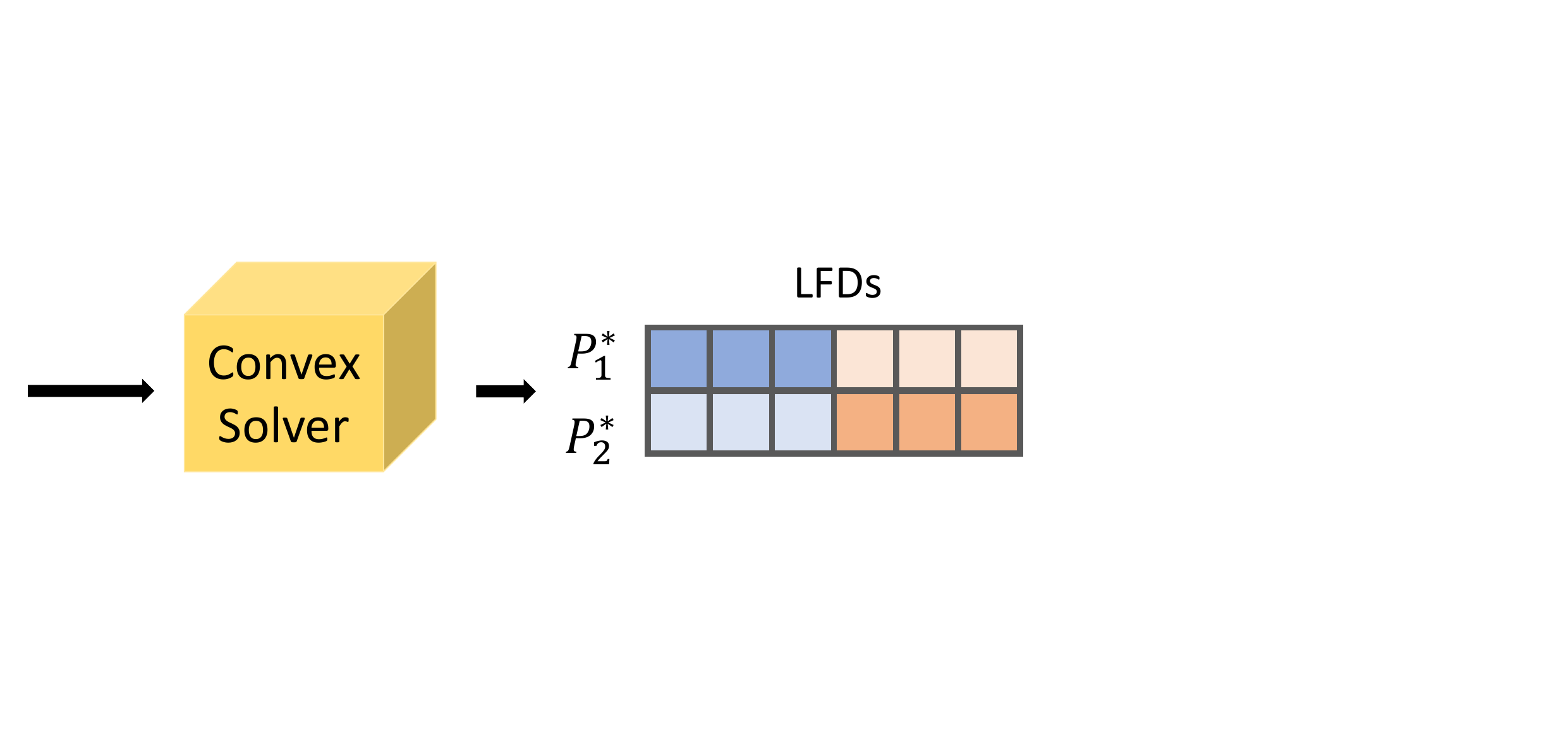}
\label{subfig:b}}
\quad 
\subfloat[Adaptive inference using optimal transport.]{\includegraphics[width=0.85\linewidth]{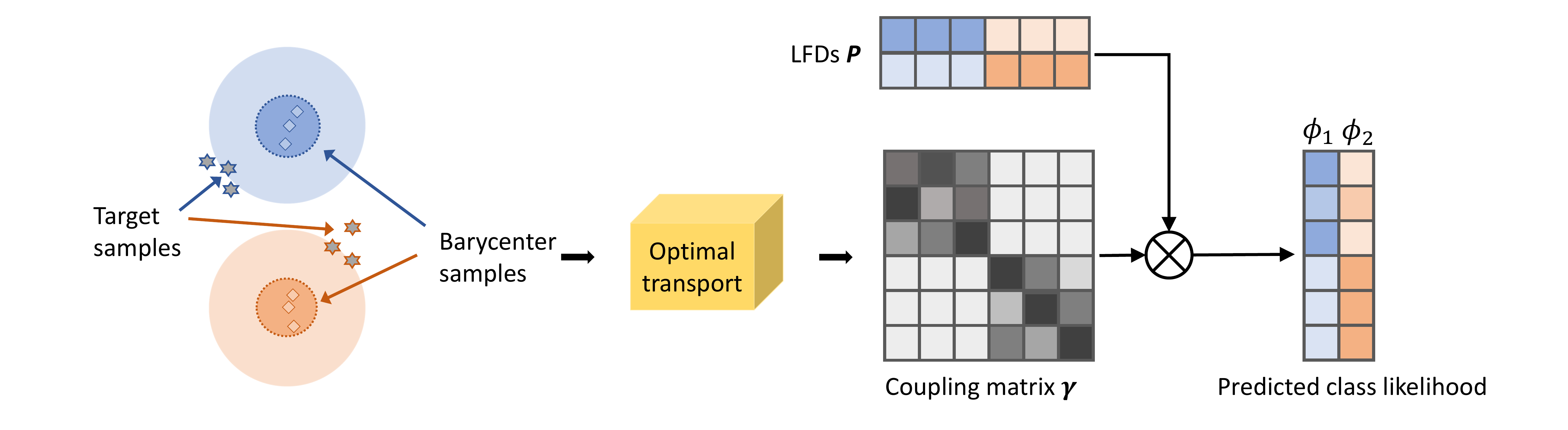}
\label{subfig:c}}
\caption{An overview of our WDRDG framework, consisting of three components: (a) Wasserstein uncertainty set construction for each class based on the empirical Wasserstein barycenters and radius obtained from given source domains. One constraint is added to control the discriminability of LFDs; (b) distributionally robust optimization to solve for the least favorable distributions; 
(c) adaptive inference for target testing samples based on probability mass on LFDs and coupling matrix from  optimal transportation between barycenter samples and target samples.
}
\label{fig_sim}
\end{figure*}

\section{Wasserstein Distributionally Robust Domain Generalization}
In this section, we present our proposed framework for domain generalization that leverages the empirical distributions from multiple source domains as shown in Figure \ref{subfig:a}, and the process of distributionally robust optimization is shown in Figure \ref{subfig:b}.
The adaptive inference for the target domain is shown in Figure \ref{subfig:c}. 
Here we show binary classification for simplicity.

More specifically, we first extrapolate the class-conditional source distributions to a Wasserstein uncertainty set for each class. Figure \ref{subfig:a} illustrates the construction of uncertainty sets of two classes. Their closeness is further controlled by the parameter $\delta$ to ensure discriminability. A convex solver then solves the distributionally robust optimization over these uncertainty sets, obtaining
the least favorable distributions (LFDs), which are represented as probability mass vectors depicted in Figure \ref{subfig:b}. Figure 
\ref{subfig:c} shows the inference process for target samples, where optimal transport \cite{courty2016optimal} is used to re-weight LFDs adaptively. 

Details of the construction of uncertainty sets and the additional Wasserstein constraints could be found in Sections \ref{method:1} and \ref{method:2}. Section \ref{method:3} discusses the re-formulation of the Wasserstein robust optimization. Adaptive inference for samples in the target domain is presented in section \ref{methodSection:adaptive}. In \ref{sec:bound}, we further analyze the generalization bound of the proposed framework.

\subsection{Construction of Uncertainty Sets}\label{method:1}
We construct the uncertainty sets controlled 
mainly by two terms: the reference distribution that represents the center of the uncertainty set,
and the radius parameter that controls the size of the set, i.e., an upper bound of the divergence between the reference distribution and other distributions in the set. 
We use {\it Wasserstein barycenter} \cite{rabin2011wasserstein}  as the reference distribution, which is the average of multiple given distributions and is capable of leveraging the inherent geometric relations among them \cite{zhou2020domain}. Given empirical class-conditional distributions $\widehat{Q}^{s_1}_k,$ $\ldots,$ $\widehat{Q}^{s_R}_k$ for each class $k$ from $R$ different source domains, the Wasserstein barycenter for class $k$ is defined as
\begin{equation} 
B^{*}_k= \underset{B_k}{\arg \min }\sum_{r=1}^{R} \frac{1}{R} \mathcal{W}_2( B_k,\widehat{Q}^{s_r}_k), k = 1,\ldots, K,
\label{eq:barycenter}
\end{equation}
which could be a proxy of the reference distribution for each uncertainty set. 
Suppose each barycenter supports on $b$ samples uniformly, i.e., $B_k= \sum_{i=1}^{b}\frac{1}{b} \delta_{\boldsymbol{x}_i^{(k)}}$, where $\{\boldsymbol{x}_i^{(k)}\}_{i=1}^{b}$ are the barycenter samples for class $k$,
then (\ref{eq:barycenter}) only optimizes over the locations $\boldsymbol{x}_i^{(k)}$.

To ensure that the uncertainty sets are large enough 
to avoid misclassification for unseen target samples, the maximum of all $R$ Wasserstein distances between class-conditional distributions of each source domain $\widehat{Q}^{s_r}_k$ and the  barycenter $B_k^*$,
is used as the radius for each class $k$:
\begin{equation}
\begin{aligned}
    \theta_k^*=\max\limits_{r=1,\ldots,R}\mathcal{W}_2\left(B^*_k, \widehat{Q}^{s_r}_k\right).
    \label{eq:radius}
\end{aligned}
\end{equation}
In this way, we can construct the Wasserstein uncertainty set $\mathcal{P}_{k}$ of radius $\theta_k^*$ centered around $B^{*}_k$ for each class $k$ following \eqref{eq:uncertainty_set}:
\begin{equation}
  \mathcal{P}_k=\left\{P_k\in \mathscr{P}(\widehat{\mathcal{X}}): \mathcal{W}_2\left(P_k, {B}_k^*\right) \leq \theta_k^*\right\}.  
\end{equation}
Figure \ref{subfig:a} shows the construction process of the uncertainty sets for two classes.

\subsection{Balance Robustness and Discriminability}\label{method:2}
\begin{figure}[!t]
\centering
\includegraphics[width = 3.5in]{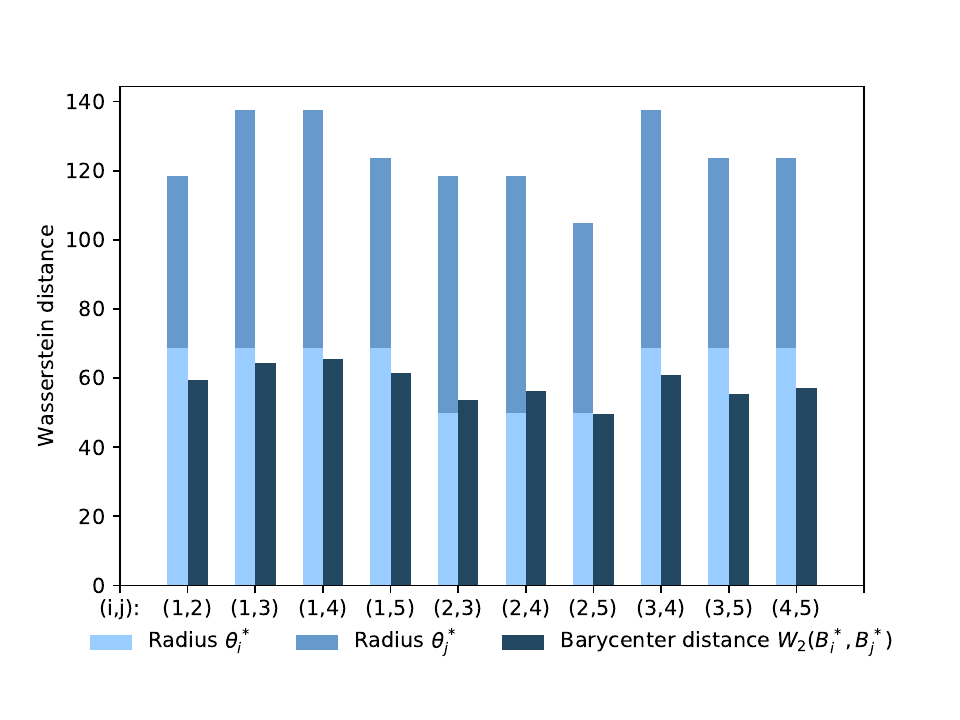}
\caption{Comparison between $\theta_i^*+\theta_j^*$ and the Wasserstein distance $W_2(B_i^*, B_j^*)$ for all $10$ unique pairs $(i,j)$ among all $5$ classes of the VLCS dataset.
The sum of uncertainty radius of any two classes is larger than the Wasserstein distance between the corresponding barycenters. The oversized radius will lead to overlapping class-specific uncertainty sets, and the distributions within them will be indistinguishable.
}
\label{fig:radius_compare}
\end{figure}

When the source training samples are limited, 
the class-conditional distributions may
vary widely in practice. 
In this situation, the radius computed from (\ref{eq:radius}) tends to be overly large,
and the uncertainty sets of different classes may overlap with each other, leading to indistinguishable LFDs for optimization problem \eqref{eq:robustHT}.
As shown in Figure \ref{fig:radius_compare}, overlap between each pair of class-specific uncertainty sets exist as the sum of their radius is larger than the Wasserstein distance between the corresponding barycenters.

Discriminability of LFDs is necessary since this leads to a well-defined problem of \eqref{eq:robustHT}, which indirectly controls the discriminability of data from different classes.
We add one more constraint to obtain significantly different LFDs that are discriminable, characterized by the Wasserstein distance between each pair of LFDs $(P_u^*, P_v^*)$ within $K$ classes:
\begin{equation}
\label{eq:wass_constraint}
\mathcal{W}_2\left(P_u^*, P_v^*\right) \ge \delta, \  1\leq u < v \leq K,
\end{equation}
where $\delta>0$ is the threshold that indicates the discriminability, which could be tuned on a validation domain. 
In this way,  robustness is ensured by large enough Wasserstein uncertainty sets, and the threshold $\delta$ guarantees discriminability among the uncertainty sets.

\subsection{Distributionally Robust Optimization}\label{method:3}
Incorporating the constraints (\ref{eq:wass_constraint}) into (\ref{eq:robustHT}), we aim to solve the following minimax problem
\begin{equation}
\label{eq:our_minimax}
\begin{aligned}
    & \min_{\phi:\mathcal{X}\rightarrow \Delta_K}  \max_{\substack{P_{k} \in \mathcal{P}_{k}, \ 1\leq k \leq K \\ 
    \mathcal{W}\left(P_u, P_v\right) \geq \delta, \ 1\leq u < v\leq K}}  \Psi\left(\phi; P_{1},\ldots, P_{K}\right).
\end{aligned}
\end{equation}
We establish the following theorem, stating a convex approximation of problem \eqref{eq:our_minimax}.

\renewcommand{\algorithmicrequire}{ \textbf{Input:}}  
\renewcommand{\algorithmicensure}{ \textbf{Output:}}    
\begin{algorithm}[t]
  \caption{Wasserstein distributionally robust domain generalization.}
  \label{algorithm}
  \begin{algorithmic}[1]
  \REQUIRE  
  $\{\widehat{Q}^{s_r}_k\}_{r=1}^R$ - empirical class-conditional distributions for each class $k$ in all $K$ classes from source domains $\{\mathcal{D}^{s_r}\}_{r=1}^R$;\\
  $b$ - number of barycenter samples for each class;\\
  $\delta$ - discriminability threshold parameter.
  \ENSURE $\phi(\boldsymbol{x}^t_j)$ - predictions for each of the unseen target samples $\{\boldsymbol{x}_j^t\}_{j=1}^{n_t}$.
  \FOR{each class $k$}
    \STATE Obtain barycenter $B^*_k$ by (\ref{eq:barycenter});\\
    \STATE Obtain radius $\theta_k^*$ using (\ref{eq:radius}).\\
    \STATE Construct uncertainty sets $\mathcal{P}_k$ centered around $B^*_k$ with radius $\theta_k^*$ as formed in (\ref{eq:uncertainty_set}).\\
  \ENDFOR
  \STATE Solve the optimization (\ref{eq:add_wass_all}) for the optimal LFDs ${P}^*_k$.\\
  \STATE 
  The inference for each target sample is given by (\ref{eq:classifier}).
  \end{algorithmic}
\end{algorithm}

\begin{theorem}
\label{theorem:add_wass_all}
Suppose the Wasserstein barycenter $B_k^\ast$ for each class as defined in (\ref{eq:barycenter}) is supported on $b$ samples. 
Let $S_b$ be the union of the support of $\{B_1^\ast,\ldots,B_K^\ast\}$ which contains ${n_b}=Kb$ samples $\{\boldsymbol{x}_i^b, i=1,\ldots,n_b\}$ in total. 
The class prior distributions of each class is denoted as $\mathbb{P}(y=k)$.
Denote each distribution within the uncertainty set $\mathcal{P}_k$ as $P_k\in\mathbb{R}_+^{n_b}$.
Let $\boldsymbol{C}\in \mathbb{R}_+^{{n_b}\times {n_b}}$ be the pairwise distance matrix of $n_b$ samples, $\boldsymbol{C}_{i,j} = \Vert \boldsymbol{x}_i^b - \boldsymbol{x}_j^b\Vert^2$, $\gamma_k\in\mathbb{R}_+^{{n_b}\times {n_b}}$ be the coupling matrix between $B_k^*$ and $P_k$, 
and $\beta_{u,v}\in\mathbb{R}_+^{{n_b}\times {n_b}}$ be the coupling matrix between any two distributions $P_u, P_v$ in different classes.
When using the Wasserstein metric of order 2, the least favorable distributions $P^*_k$ of the problem (\ref{eq:our_minimax}) could be obtained by solving: 
\begin{equation}
\label{eq:add_wass_all}
\begin{aligned}
\max_{\substack{P_1,\ldots,P_{K}\in\mathbb{R}_+^{{n_b}}\\\gamma_1,\ldots,\gamma_{K}\in\mathbb{R}_+^{{n_b}\times {n_b}}\\ \beta_{u,v}\in\mathbb{R}_+^{{n_b}\times {n_b}}}}
  & 1-\sum_{i=1}^{{n_b}}\max_{1\leq k \leq K}\mathbb{P}(y=k) P_k\left(\boldsymbol{x}^b_i\right)\\
  \text { s.t. } 
  & \left\langle{\gamma}_k,\boldsymbol{C}\right\rangle_F\leq (\theta_k^*)^2,\left\langle{\beta}_{u,v},\boldsymbol{C}\right\rangle_F\geq \delta^2,\\
  & {\gamma}_k \boldsymbol{1}_{{n_b}}=B_k^*,{\gamma}_k^T \boldsymbol{1}_{{n_b}}=P_k,\\
  & {\beta}_{u,v} \boldsymbol{1}_{{n_b}}=P_u,{\beta}_{u,v}^T \boldsymbol{1}_{{n_b}}=P_v,\\
  & \forall 1 \leq k \leq K, 1\leq u< v \leq K,
\end{aligned}
\end{equation}
and the optimal prediction function of (\ref{eq:our_minimax}) satisfies
$\phi^*_k(\boldsymbol{x}_i^b)={P_{k}^{*}\left(\boldsymbol{x}_i^b\right)}/{\sum_{k=1}^{K}P_{k}^{*}\left(\boldsymbol{x}_i^b\right)}$ for any $\boldsymbol{x}_i^b\in S_b$.
\end{theorem}
\noindent
The constraints on $\gamma_k$ restrict each target class-conditional distribution to its respective uncertainty set of radius $\theta_k^*$.
The constraints on $\beta_{u,v}$ restrict the Wasserstein distance between each pair of class-conditional distributions in the target domain following \eqref{eq:wass_constraint}.
Based on the above theorem,
the classification for any sample in the sample set $S_b$ is given by $\Phi(\boldsymbol{x}_i^b)=\arg\max_{1\leq k\leq K}P^*_k(\boldsymbol{x}_i^b)$.
{The proof can be found in 
the supplementary material.}

\subsection{Adaptive Inference by Test-time  Adaptation}

\label{methodSection:adaptive}
Since the barycenters are the weighted average of distributions from multiple source domains, the barycenter samples in the support set $S_b$ could be viewed as samples from a {\it generalized source domain} denoted as $\mathcal{D}^{b}$. 
For any sample in $\mathcal{D}^{b}$, the likelihood that it is assigned to each class could be decided based on $\phi^*$ 
by a non-parametric  inference method such as KNN \cite{zhu2020distributionally}. 
When making predictions for samples from an unseen target domain $D^t$, the domain shift between $\mathcal{D}^{b}$ and $D^t$ needs to be considered. We adopt optimal transport   to reduce the domain shift adaptively by the following test-time adaptation process.

Suppose $\widehat{\mu}_{b}=\sum_{i=1}^{n_{b}}\frac{1}{n_{b}} \delta_{\boldsymbol{x}_i^{b}}$ and $\widehat{\mu}_{t}=\sum_{j=1}^{n_t}\frac{1}{n_t} \delta_{\boldsymbol{x}_j^{t}}$ are the empirical marginal distributions of the feature vectors from the generalized source domain $\mathcal{D}^{b}$ and a target domain $\mathcal{D}^t$, respectively.
Denote the coupling matrix of transporting from target to the generalized source distribution using optimal transport \cite{courty2016optimal}
as $\boldsymbol{\gamma}=[{\gamma}_1,\ldots,{\gamma}_{n_t}]^T\in\mathbb{R}^{n_t \times n_b}$, where each vector ${\gamma}_{j}\in\mathbb{R}^{n_b}$, $j=1,\ldots,n_t$, represents the transported mass from
the $j$-th target sample to each of the $n_b$ barycenter samples.
In most optimal transport-based domain adaptation methods, each target sample $\boldsymbol{x}_j^{t}$, $j=1,\ldots, n_t$, is first transported to $\widehat{\boldsymbol{x}}_j^{t}$ in the generalized source domain $\mathcal{D}^{b}$ by the barycentric mapping:
\begin{equation}
    \widehat{\boldsymbol{x}}_j^{t}=\sum_{i=1}^{n_b}n_t {\boldsymbol{\gamma}}_{j,i}{\boldsymbol{x}_i^{b}}, \ j=1,\ldots, n_t,\label{barycentric_mapping}
\end{equation}
then having its label inferred based on the classifier learned on the labeled samples.
Instead of such a two-step process, we propose an equivalent single-step inference process. The following proposition states the equivalence,
and the proof can be found in the supplementary.

\begin{proposition}
\label{proposition: equivalent}
Given the coupling matrix $\boldsymbol{\gamma}\in\mathbb{R}^{n_t \times n_b}$. Suppose we transport the target sample $\boldsymbol{x}_j^{t}$ from the empirical target distribution $\widehat{\mu}_{t}=\sum_{j=1}^{n_t}\frac{1}{n_t} \delta_{\boldsymbol{x}_j^{t}}$ to the generalized source domain empirical distribution $\widehat{\mu}_{b}=\sum_{i=1}^{n_{b}}\frac{1}{n_{b}} \delta_{\boldsymbol{x}_i^{b}}$ 
by the barycentric mapping as shown in (\ref{barycentric_mapping}), and obtain the class likelihood by re-weighting $\phi^*_k(\boldsymbol{x}_i^b)$ of all the samples $\boldsymbol{x}_i^b \in S_b$ using the weight function $w\left(\widehat{\boldsymbol{x}}_j^{t},\boldsymbol{x}_i^{b}\right)=n_t \boldsymbol{\gamma}_{j,i}$.
Then the resulting classifier is equivalent to directly re-weighting LFDs on the barycenter samples using the coupling matrix. The equivalent classification result is:
\begin{equation}
    \Phi(\boldsymbol{x}^t_j)=\underset{1 \leq k \leq K}{\arg \max }\sum_{i=1}^{n_b}{\boldsymbol{\gamma}_{j,i}} P^*_k(\boldsymbol{x}^{b}_{i}).
\end{equation}
\end{proposition}
This proposition illustrates that domain difference between target domain and generalized source domain can be eliminated by adaptively  applying the coupling matrix in the inference stage, without actually transporting the target samples to the generalized source domain.


Denote the LFDs for all classes as $\boldsymbol{P}=\left[{P}_1^*,\ldots,{P}_K^*\right]^T\in\mathbb{R}^{K\times n_b}$.
Based on Proposition \ref{proposition: equivalent},
the predicted class likelihood of each target sample $\boldsymbol{x}^t_j$ can be written as
\begin{equation}
\label{eq:classifier}
\phi(\boldsymbol{x}^t_j)=\frac{ {{\boldsymbol{\gamma}_j}}^T \boldsymbol{P}^T}{ {{\boldsymbol{\gamma}_j}}^{T} \boldsymbol{P}^{T}\boldsymbol{1}_{K}}=\left[\phi_1(\boldsymbol{x}^t_j), \ldots, \phi_{K}\left(\boldsymbol{x}^t_j\right)\right],
\end{equation}
where $0\leq \phi_k\left(\boldsymbol{x}^t_j\right)\leq 1, \sum_{k=1}^{K} \phi_k\left(\boldsymbol{x}^t_j\right) = 1$.
The algorithm is summarized in Algorithm \ref{algorithm}.
Further adding the optimal-transport based adaptive inference leads to our complete framework Wasserstein \textbf{D}istributionally \textbf{R}obust \textbf{D}omain \textbf{G}eneralization (\textbf{WDRDG}).

\subsection{Generalization Analysis}\label{sec:bound}

We further analyze the generalization risk of our proposed method.
Our analysis
considers the domain shift between the target domain and the generalized source domain. 

Based on (\ref{eq:classifier}), the classification decision for the test sample $\widehat{\boldsymbol{x}}_j^{t}$ in the target domain is based on the weighted average
\begin{equation}
\label{eq:argmax_reweight}
     \underset{1 \leq k \leq K}{\arg \max }\sum_{i=1}^{n_b} w\left(\widehat{\boldsymbol{x}}_j^{t},\boldsymbol{x}_i^{b}\right) P^*_k(\boldsymbol{x}^{b}_{i}).
\end{equation}

Consider a binary classification problem with label set $\{0,1\}$.
Let $\phi(\boldsymbol{x})=[\phi_0(\boldsymbol{x}),\phi_1(\boldsymbol{x})]$ represents the prediction vector of $\boldsymbol{x}$ belonging to either classes.
The true labeling function is denoted as $f:\mathcal{X}\rightarrow\{0,1\}$.
Considering the simple case that all classes are balanced, 
the expected risk that the correct label is not accepted for samples in any distribution $\mu$ is denoted as $\epsilon_{\mu}\left(\phi\right)=\mathbb{E}_{\boldsymbol{x}\sim\mu}[1-\phi_{f(\boldsymbol{x})}(\boldsymbol{x})]$.
We now present the following theorem stating the generalization bound.

\begin{theorem}
\label{theorem:bound}
Suppose the distributionally robust prediction function $\phi^{S_b}$ learned from the sample set $S_b$ is $M$-Lipschitz continuous for some $M\geq0$.
Let $\mu_b$ and $\mu_t$ be the probability distributions for the generalized source and target domain, respectively.
Then the risk on the target distribution $\mu_t$ follows
\begin{equation}\label{eq:generalization}
\begin{aligned}
\epsilon_{\mu_{t}}(\phi^{S_b}) 
&\leq \epsilon_{\mu_{b}}(\phi^{S_b}) + 2M \cdot \mathcal{W}_{1}\left(\mu_{b}, \mu_{t}\right)+\lambda,
\end{aligned}
\end{equation}
where $\lambda=\underset{\phi:\mathcal{X}\rightarrow [0,1], \Vert\phi\Vert_{\textrm{Lip}}\leq M}{\min}\left(\epsilon_{\mu_t}(\phi) + \epsilon_{\mu_b}(\phi)\right)$.
\end{theorem}
\noindent
The first term is the risk on the barycenter distribution $\mu_b$.
The second term shows that the divergence between the barycenter distribution and target distribution, measured by the Wasserstein distance (of order $1$).
This theorem shows that the generalization risk on the target domain is affected by
the Wasserstein distance between the barycenter distribution and the target distribution, which represents the gap between the generalized source domain and the target domain.

By applying the concentration property of the Wasserstein distance \cite{bolley2007quantitative}, we can measure the generalization risk based on empirical Wasserstein distances similar to Theorem 3 in \cite{shen2018wasserstein}. Under the assumption of Theorem \ref{theorem:bound}, if the two probability distributions $\mu_b$ and $\mu_t$ satisfy $T_1(\xi)$ inequality \cite{bolley2007quantitative}, 
then for any $d^\prime > d$ and $\xi^\prime < \xi$, there exists some constant $N_0$ depending on $d^\prime$ such that for any $\varepsilon>0 $ and $\min (n_b, n_t) \geq N_{0} \max \left(\varepsilon^{-\left(d^{\prime}+2\right)}, 1\right)$,
with probability at least $1-\varepsilon$ the following holds for the risk on the target domain
\[
\begin{aligned}
\epsilon_{\mu_t}(\phi^{S_b}) \leq &  \epsilon_{\mu_b}(\phi^{S_b})
  +2M \mathcal{W}_{1}\left(\widehat{\mu}_{b}, \widehat{\mu}_{t}\right)+\lambda\\
  &
  +2 M \sqrt{2 \log \left(\frac{1}{\varepsilon}\right) / \xi^{\prime}}\left(\sqrt{\frac{1}{n_{b}}}+\sqrt{\frac{1}{n_{t}}}\right).\notag
\end{aligned}
\]
Here $d$ denotes the dimension of the feature space. The last term illustrates the importance of getting more labeled samples from the generalized source domain.
This result show that reducing the Wasserstein distance between the barycenters and target distributions will lead to tighter upper bound for the risk of the learned model on the target domain.
Therefore, it provides a theoretical motivation to our design of the test-time adaptation, which reduces such domain gap by optimal transport.
Details of the proof could be found in the supplementary material.

\section{Experiments}

\subsection{Datasets}
To evaluate the effectiveness of our proposed domain generalization framework,
we conduct experiments on three datasets: 
the VLCS \cite{fang2013unbiased} dataset, the PACS \cite{li2017deeper} dataset, and the Rotated MNIST \cite{ghifary2015domain} dataset.

\noindent
\textbf{VLCS dataset}
This domain generalization benchmark contains images from four image classification datasets: PASCAL VOC2007 (V), LabelMe (L), Caltech-101 (C), and SUN09 (S),
denoted as domains $D_V$, $D_L$, $D_C$, and $D_S$, respectively \cite{torralba2011unbiased}. 
There are five common categories: bird, car, chair, dog and person.

\noindent
\textbf{PACS dataset}
The PACS dataset contains images of four domains: Photos (P), Art painting (A), Cartoon (C) and Sketch (S) \cite{li2017deeper}.
There are in total $7$ types of objects in this classification task, i.e., dog, elephant, giraffe, guitar, horse, house, and person. 

\begin{figure}
  \centering
  \subfloat[$r_0$]
    {\includegraphics[width=0.3\linewidth]{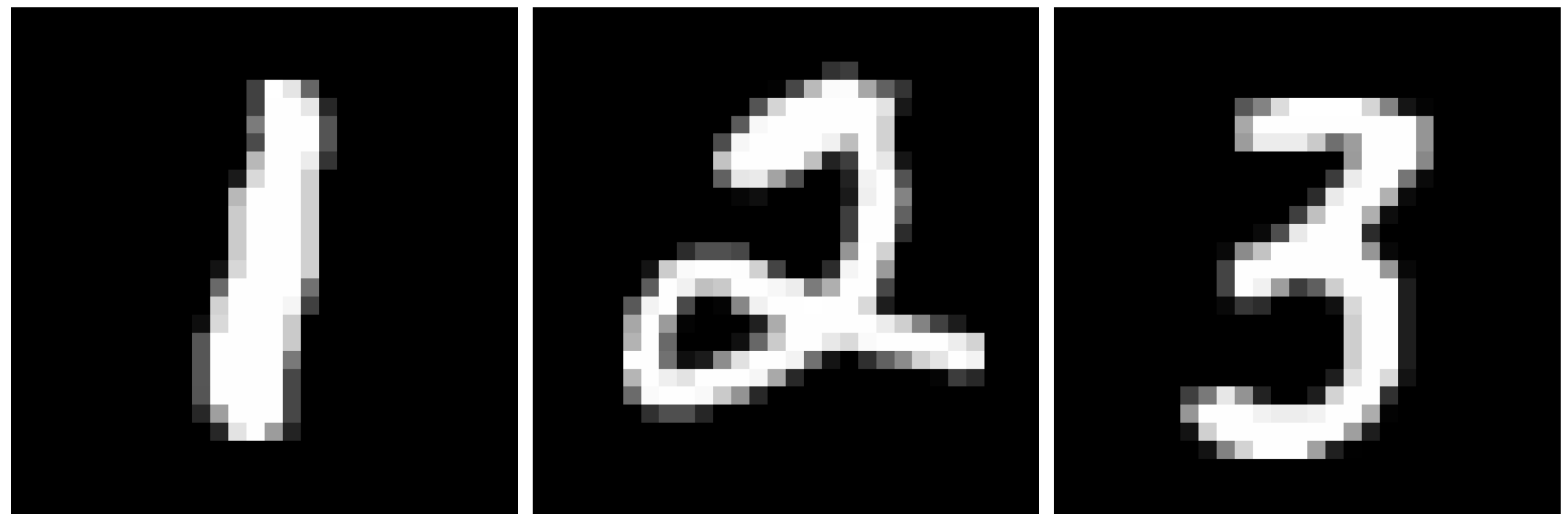}\label{subfig:rmnist_images_0}}
  \subfloat[$r_{30}$]
    {\includegraphics[width=0.3\linewidth]{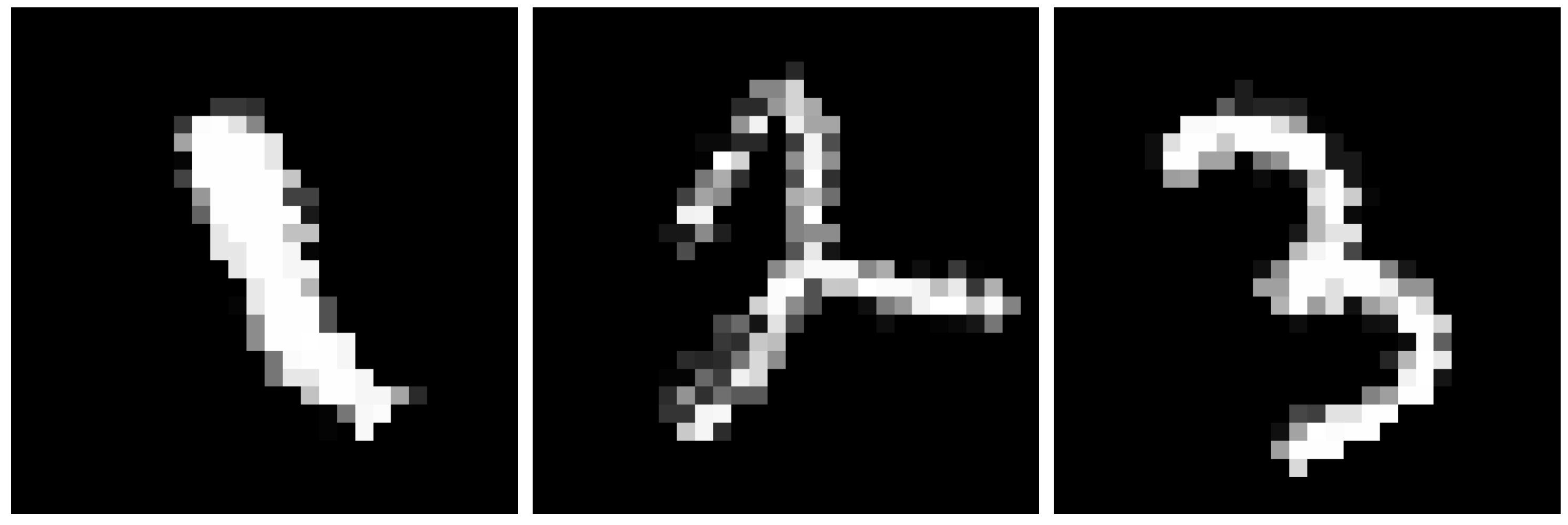} \label{subfig:rmnist_images_30}}
  \quad 
  \subfloat[$r_{60}$]
    {\includegraphics[width=0.3\linewidth]{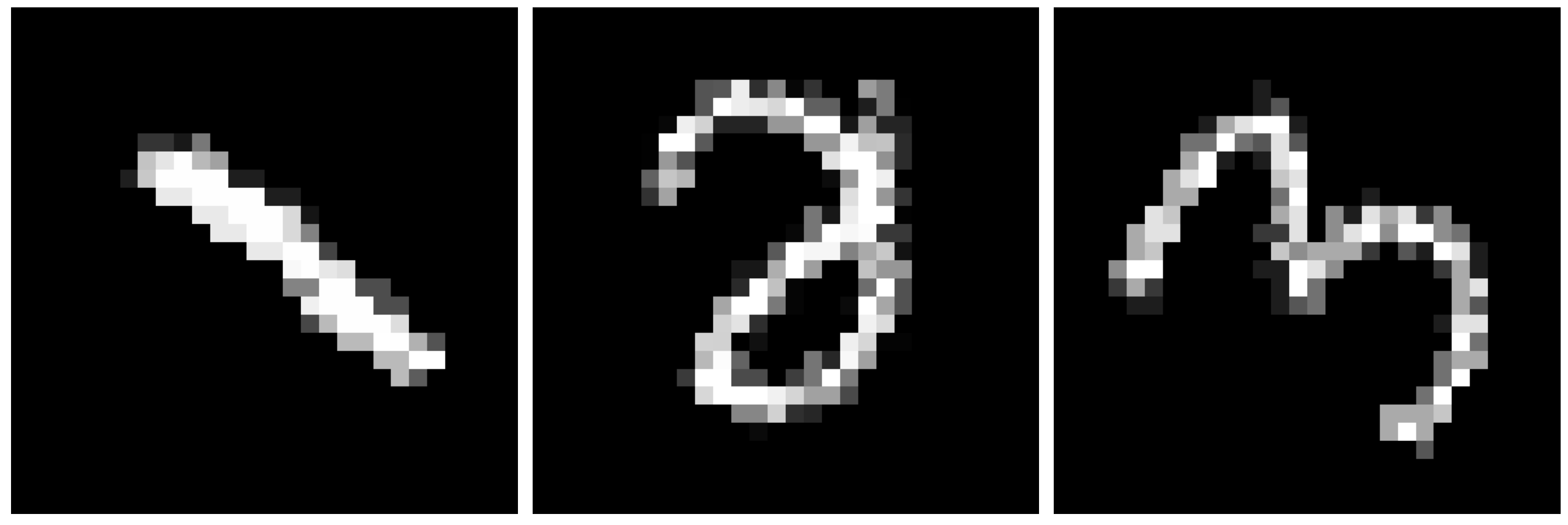} \label{subfig:rmnist_images_60}}
  \subfloat[$r_{90}$]
    {\includegraphics[width=0.3\linewidth]{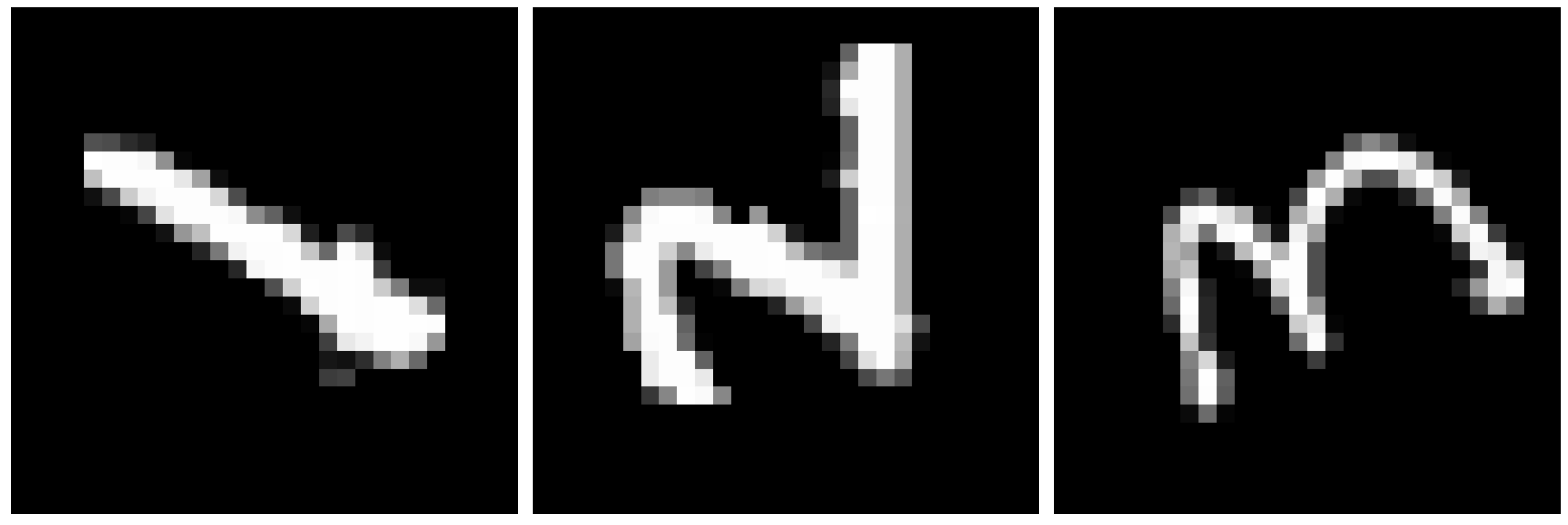} \label{subfig:rmnist_images_90}}
  \caption{Visualization of example images from four domains of the Rotated MNIST dataset with rotation angles of $0^\circ$, $30^\circ$, $60^\circ$, $90^\circ$.}
  \label{fig:rmnist_images}
\end{figure}

\noindent
\textbf{Rotated MNIST dataset}
We constructed the Rotated MNIST dataset with four domains, $r_0$, $r_{30}$, $r_{60}$ and $r_{90}$ following the common settings \cite{ghifary2015domain}.
$r_0$ denotes the domain containing original images from the MNIST dataset, and
we rotated each image in the original MNIST dataset by $30$, $60$ and $90$ degrees clockwise, respectively to generate the dataset of $r_{30}$, $r_{60}$ and $r_{90}$.
Some example images are shown in Figure \ref{fig:rmnist_images}.
We randomly sampled 
among digits $[1, 2, 3]$.

\subsection{Experimental Configuration}
We evaluate each method on the multi-domain datasets via the leave-one-domain-out experiments,
i.e., we train a model based on the source domains and test on the hold-out unseen target domain.
For example, when the target domain is $D_V$, then the transfer direction is from three source domains to a target domain, i.e., $D_L, D_C, D_S\rightarrow D_V$, and the average of test accuracies of four cross-domain experiments is taken as the final result.

We mainly consider the scenario when we have only limited labeled data from the source domains.
Therefore, for each domain, we randomly select some images to form the training set, validation set and test set for the cross-domain classification. 
The training set is used to learn robust models, whose
parameters are then selected on the validation set aggregated by the validation sets of each source domain.
The performance of a model is finally evaluated on the test set.
Details of the sets for training, validation and testing are as follows:
\begin{itemize}
    \item \textbf{Training set} For each domain, we randomly select up to $25$ images. To be more specific, we set the number of training images per category per domain to be a number in the set $\{2,3,5,7,10, 15, 20, 25\}$. 
    The training data from the three source domains form the training set.
    \item \textbf{Validation set} For each domain, $10$ images per category are randomly selected. The validation data from the three source domains form the validation set.
    \item \textbf{Test set} We sample $20$ images per category for each domain. The sampled test data from the unseen target domain form the test set.   
\end{itemize}
We repeat the above sampling process $5$ times for all datasets, so that the experiments are based on $5$ trials.The average results of all $5$ trials are finally reported.

Features pretrained on neural networks are taken as our input.
For the Rotated MNIST dataset, the Resnet-18 \cite{he2016deep} pretrained on the ImageNet is used to extract $512$-dimensional features as the inputs.
For the VLCS dataset, the pretrained $4096$-dimensional DeCAF features \cite{donahue2014decaf} are employed as the inputs of our algorithm following previous works \cite{motiian2017unified, dou2019domain}.
For the PACS dataset, we use the ImageNet pre-trained AlexNet \cite{krizhevsky2012imagenet} as the backbone network to extract the $9216$-dimensional features.


\subsection{Baseline Methods}
We compare our proposed WDRDG framework with the following baseline methods in terms of the average classification accuracy.
All methods for comparison are summarized as below:
\begin{itemize}
    \item {KNN:} We adopt the combination of training instances from all source domains to train the nearest neighbor classifier.
    \item {MDA \cite{hu2020domain}:} We apply Multidomain Discriminant Analysis (MDA) to learn domain-invariant feature transformation
    that is applicable when $P(X|Y)$ changes across domains.
    1-NN is adopted as a classifier to the learned feature transformations for classification.
    \item {CIDG \cite{li2018domain2}:} Conditional Invariant Domain Generalization (CIDG) finds a linear transformation to minimize the total domain scatter with regard to each class-conditional distributions. The learned features are also classified using KNN.
    \item {MLDG \cite{li2018learning}}: We consider this meta-learning based domain generalization method as another baseline which models unseen target domain shift. 
    A simple two-layer network is trained to learn the classifier.
    
\end{itemize}

For our proposed WDRDG framework, 
we use the CVXPY package \cite{diamond2016cvxpy} to solve the distributionally robust optimization problem.
The Wasserstein distance of order 2 is used for all experiments, and calculated with the POT package \cite{flamary2021pot}.

\subsection{Results and Discussion}
\begin{figure*}[!t]
  \centering
  {\includegraphics[width=0.23\linewidth]{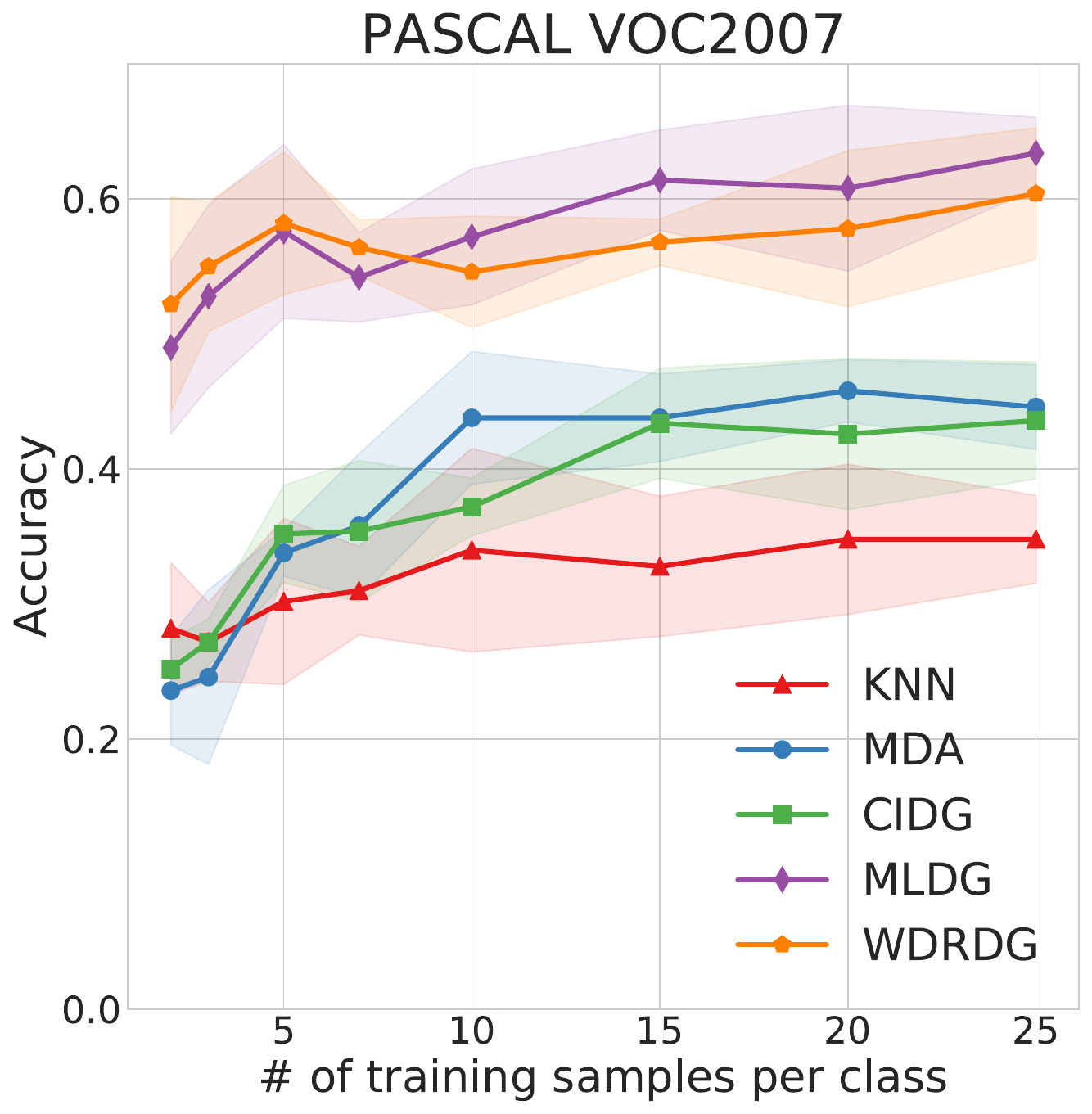}\label{subfig:vlcs_results_V}}
  {\includegraphics[width=0.23\linewidth]{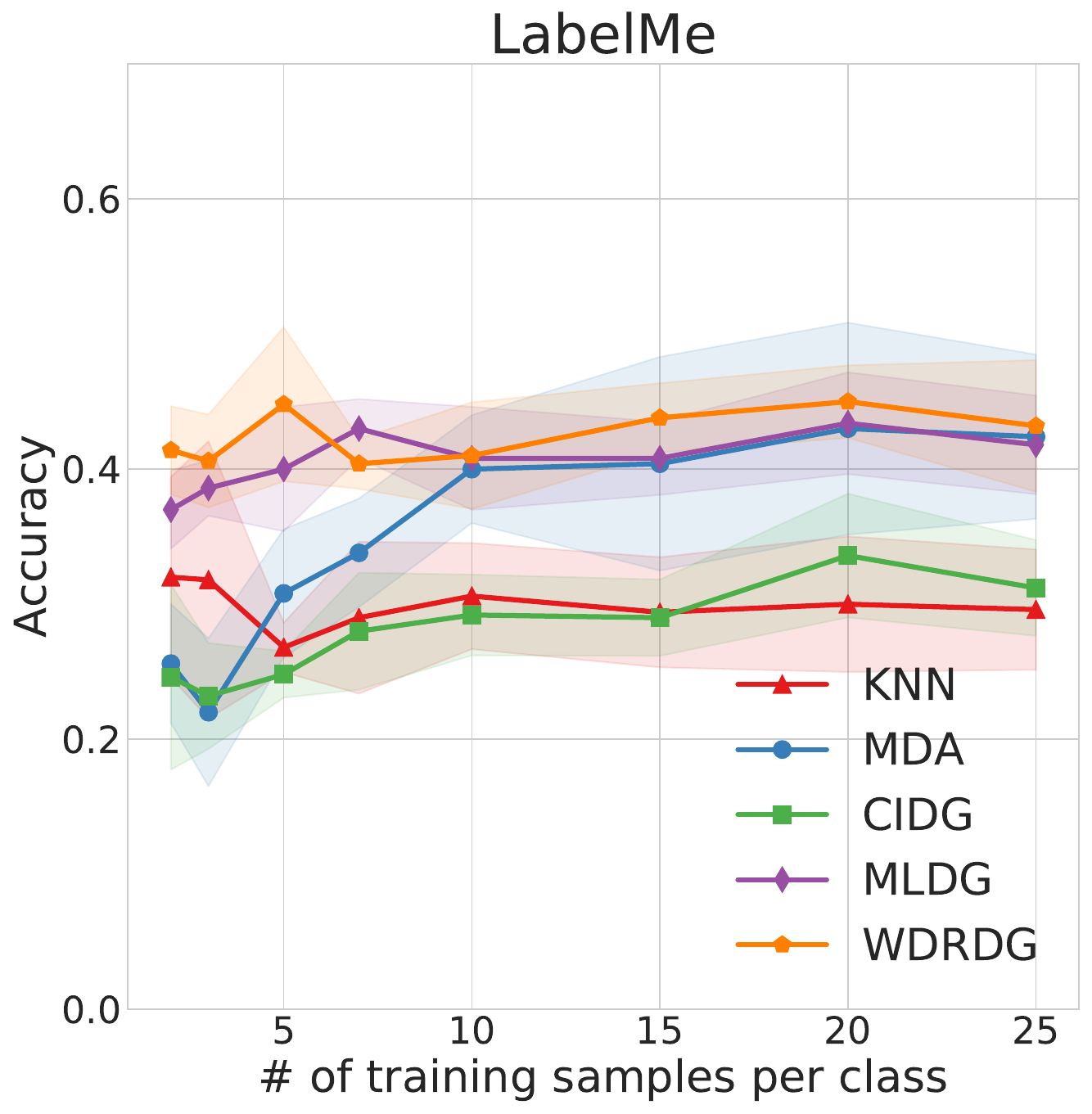}\label{subfig:vlcs_results_L}}
  {\includegraphics[width=0.23\linewidth]{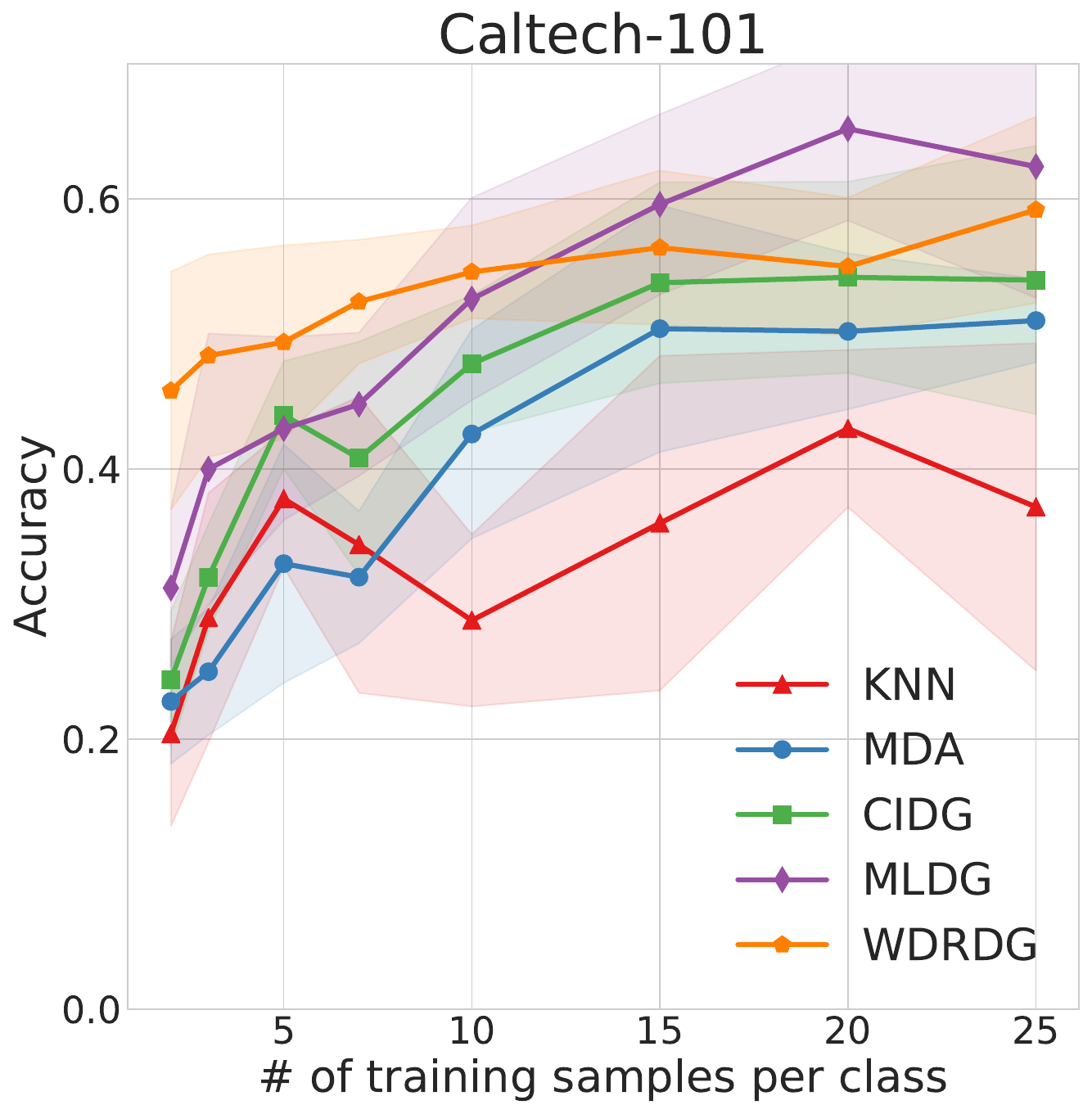}\label{subfig:vlcs_results_C}}
  {\includegraphics[width=0.23\linewidth]{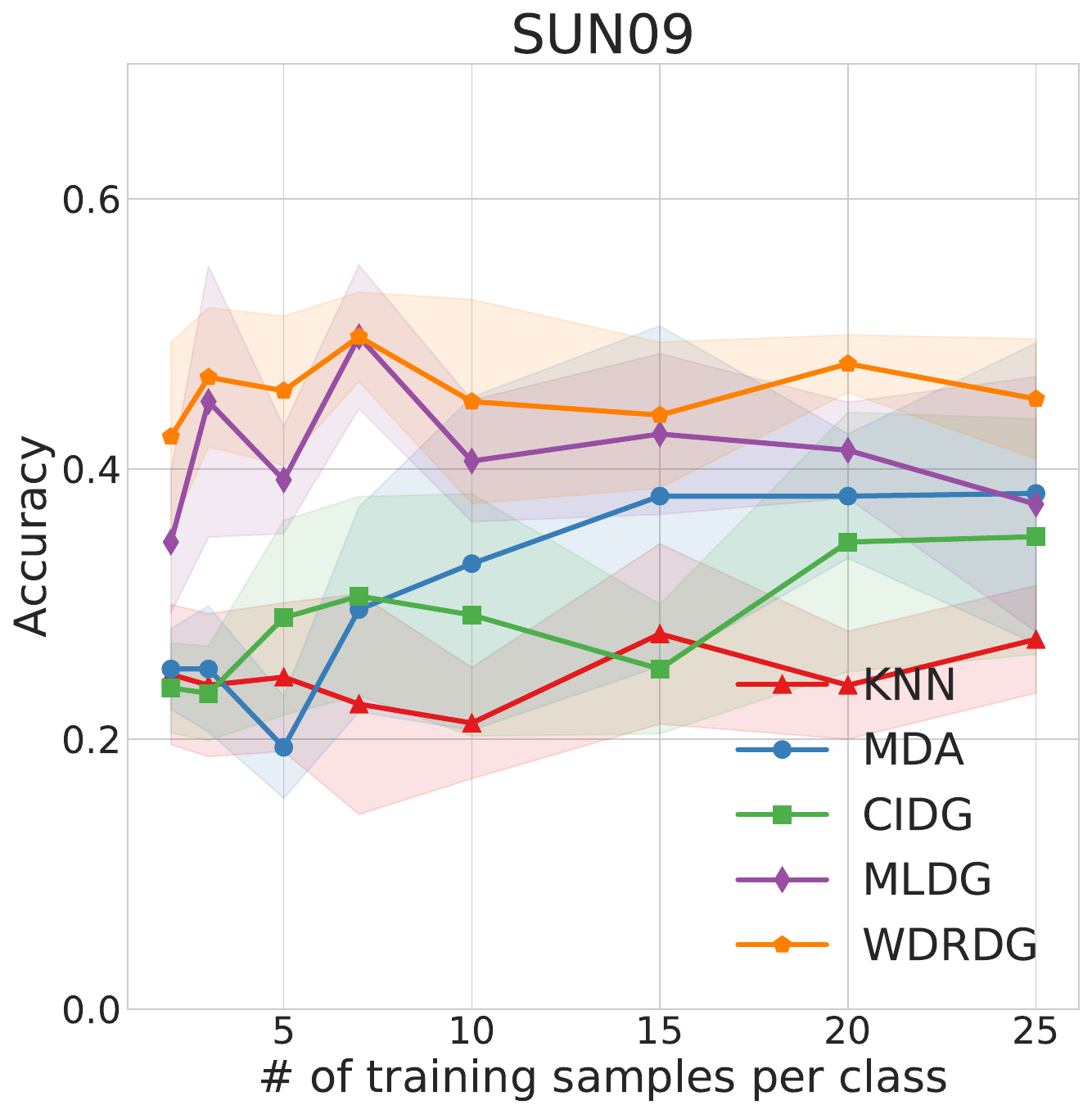}\label{subfig:vlcs_results_S}}
  \quad 
  {\includegraphics[width=0.23\linewidth]{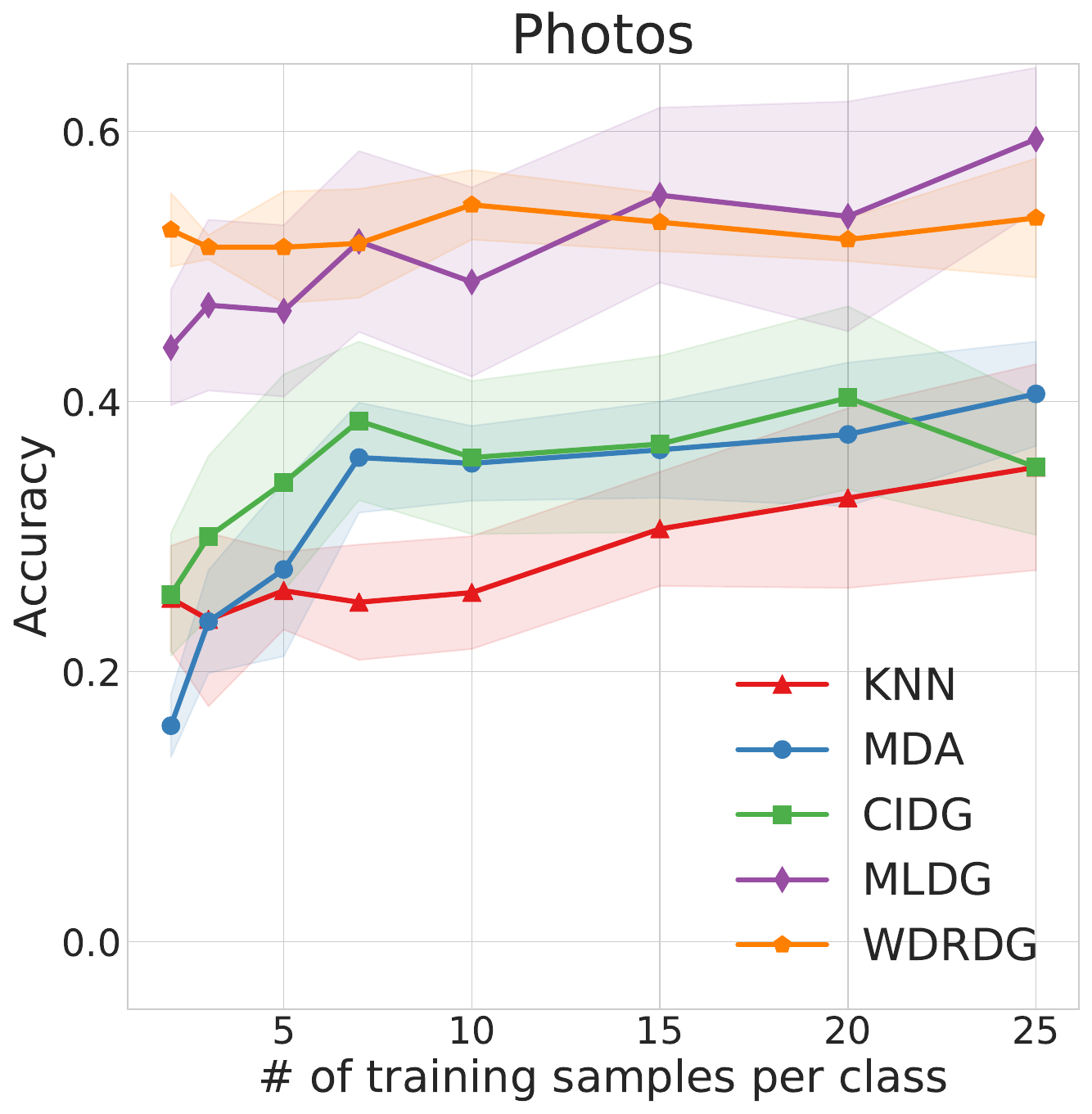}\label{subfig:pacs_results_P}}
  {\includegraphics[width=0.23\linewidth]{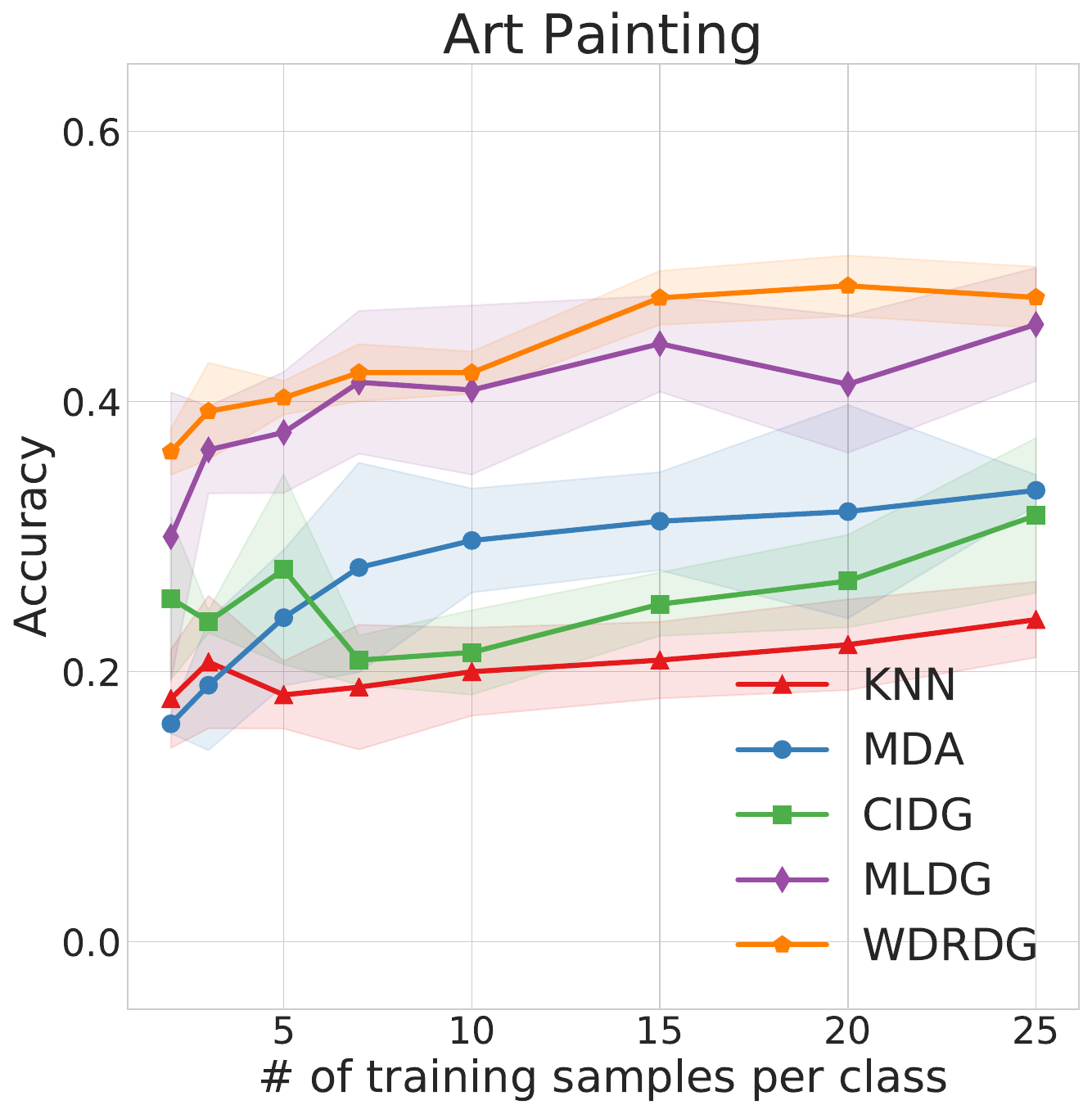}\label{subfig:pacs_results_A}}
  {\includegraphics[width=0.23\linewidth]{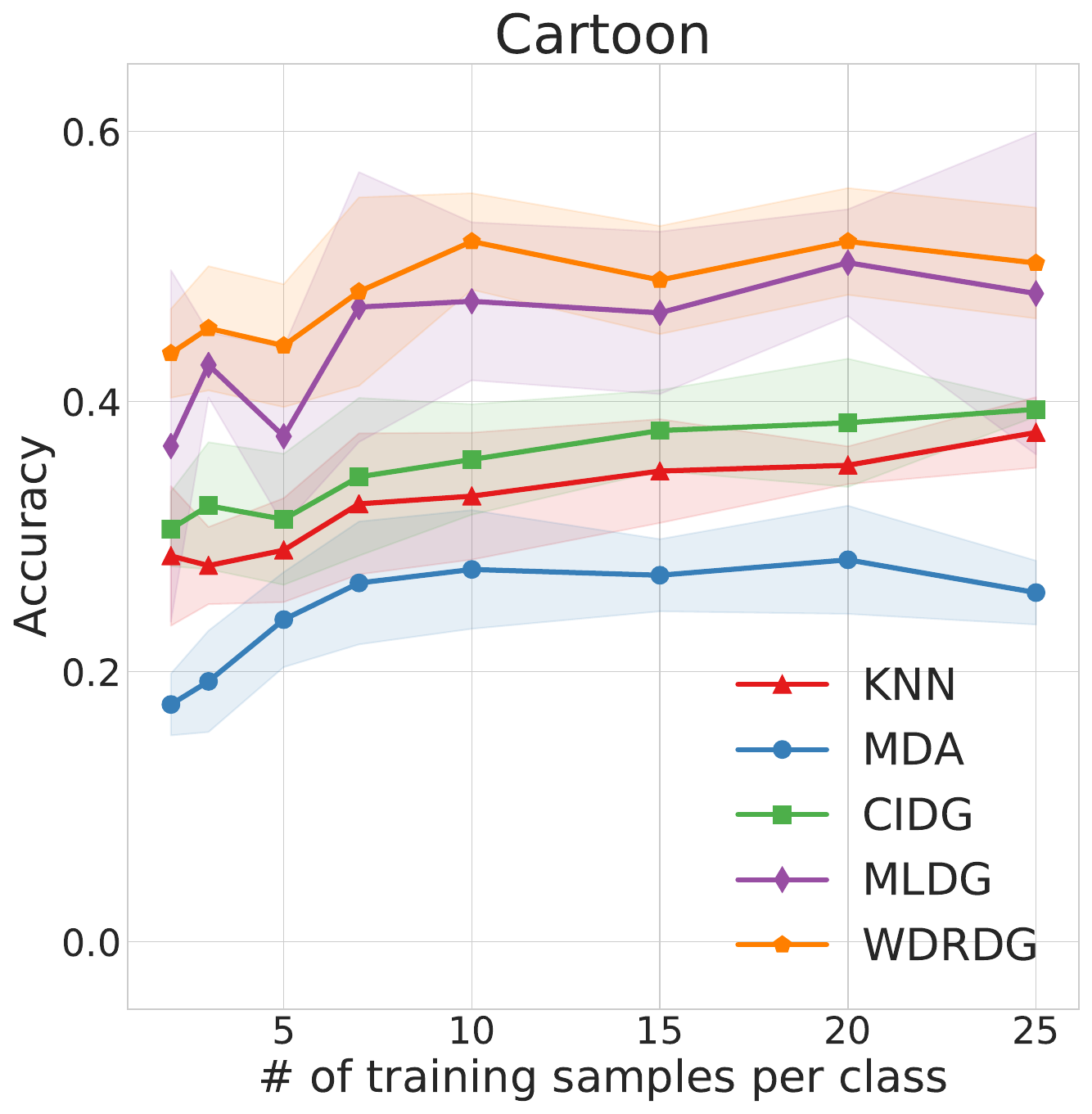}\label{subfig:pacs_results_C}}
  {\includegraphics[width=0.23\linewidth]{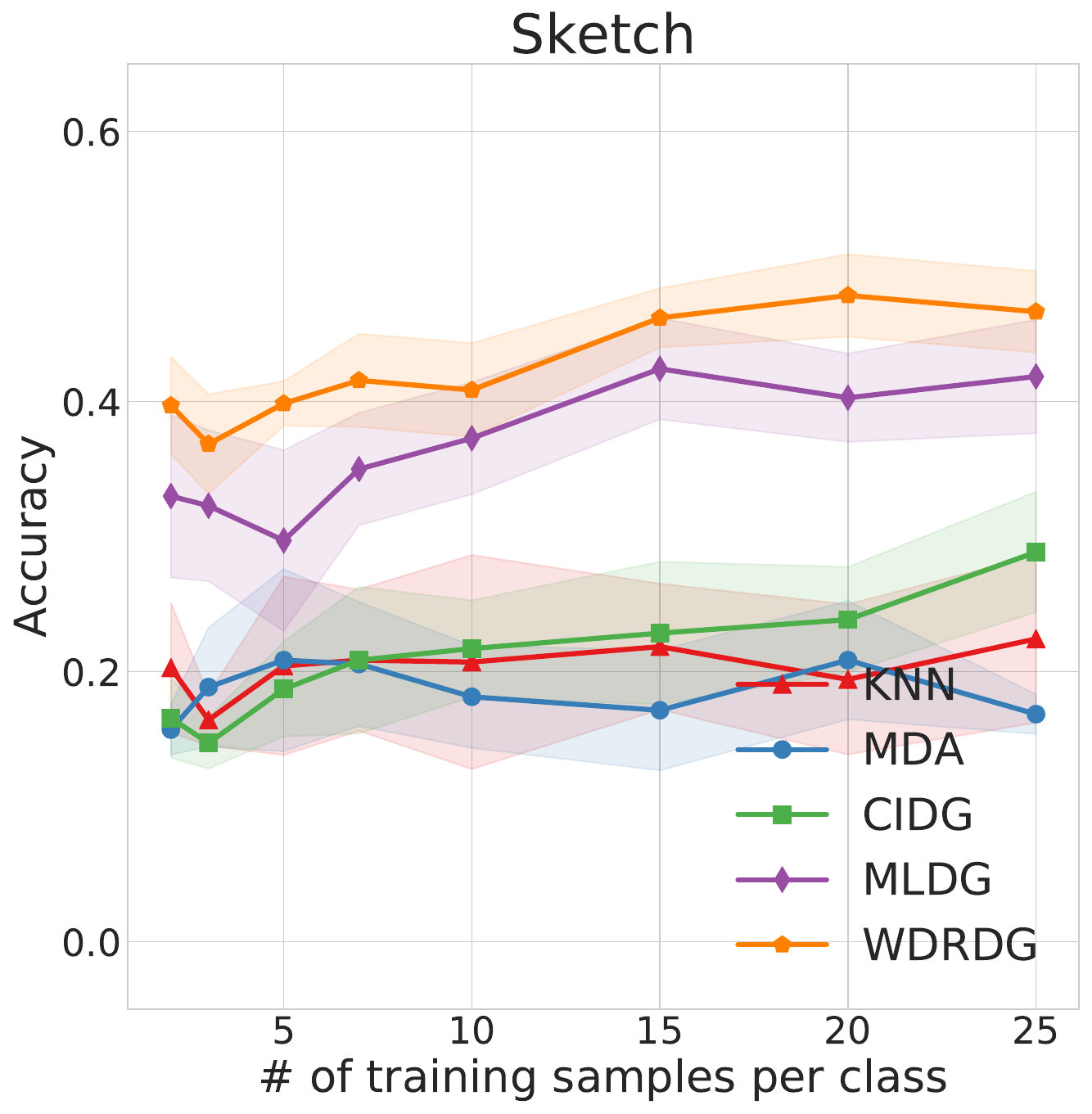}\label{subfig:pacs_results_S}}
  \quad 
  {\includegraphics[width=0.23\linewidth]{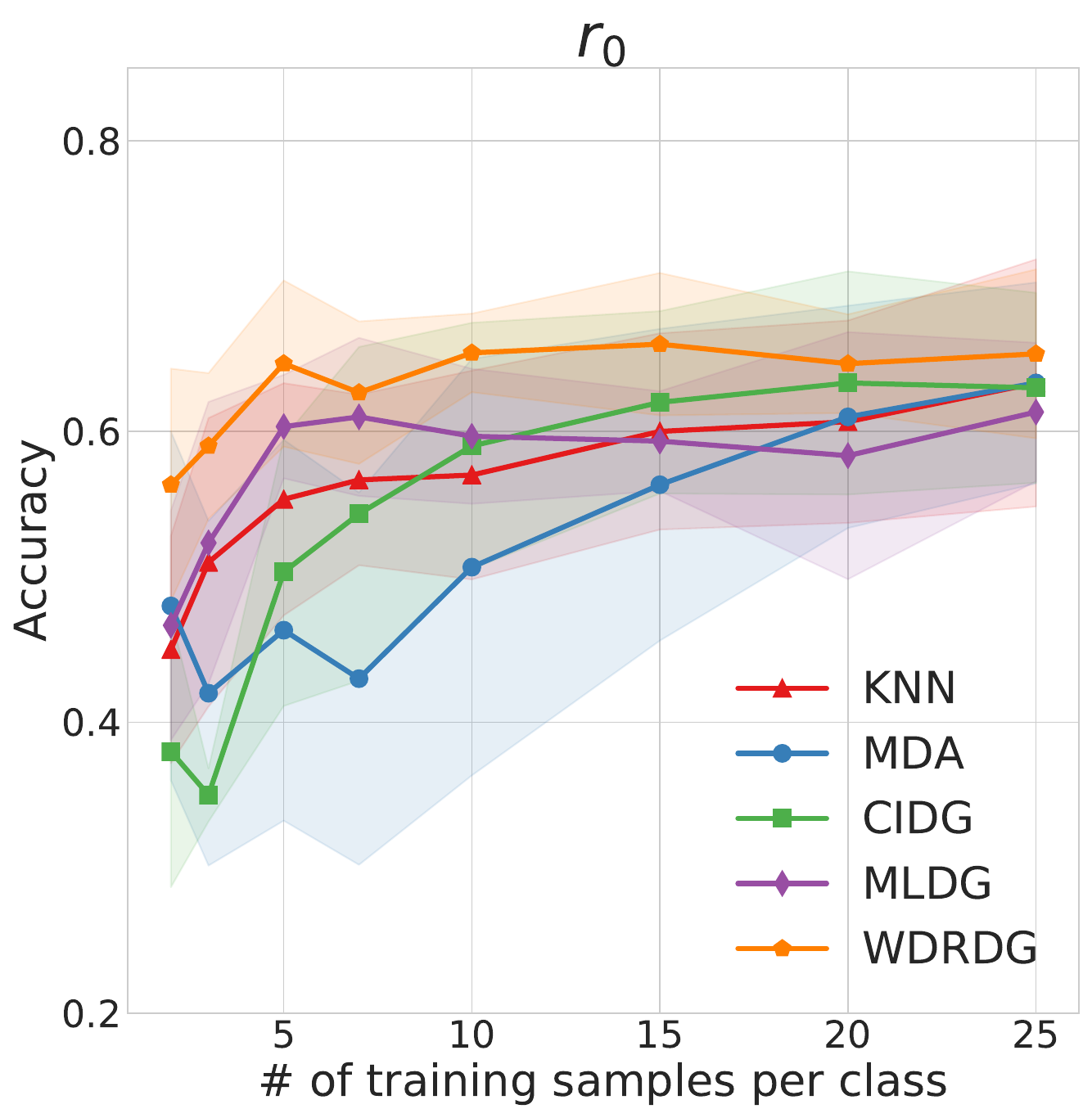}  \label{fig:rmnist_results_0}}
  {\includegraphics[width=0.23\linewidth]{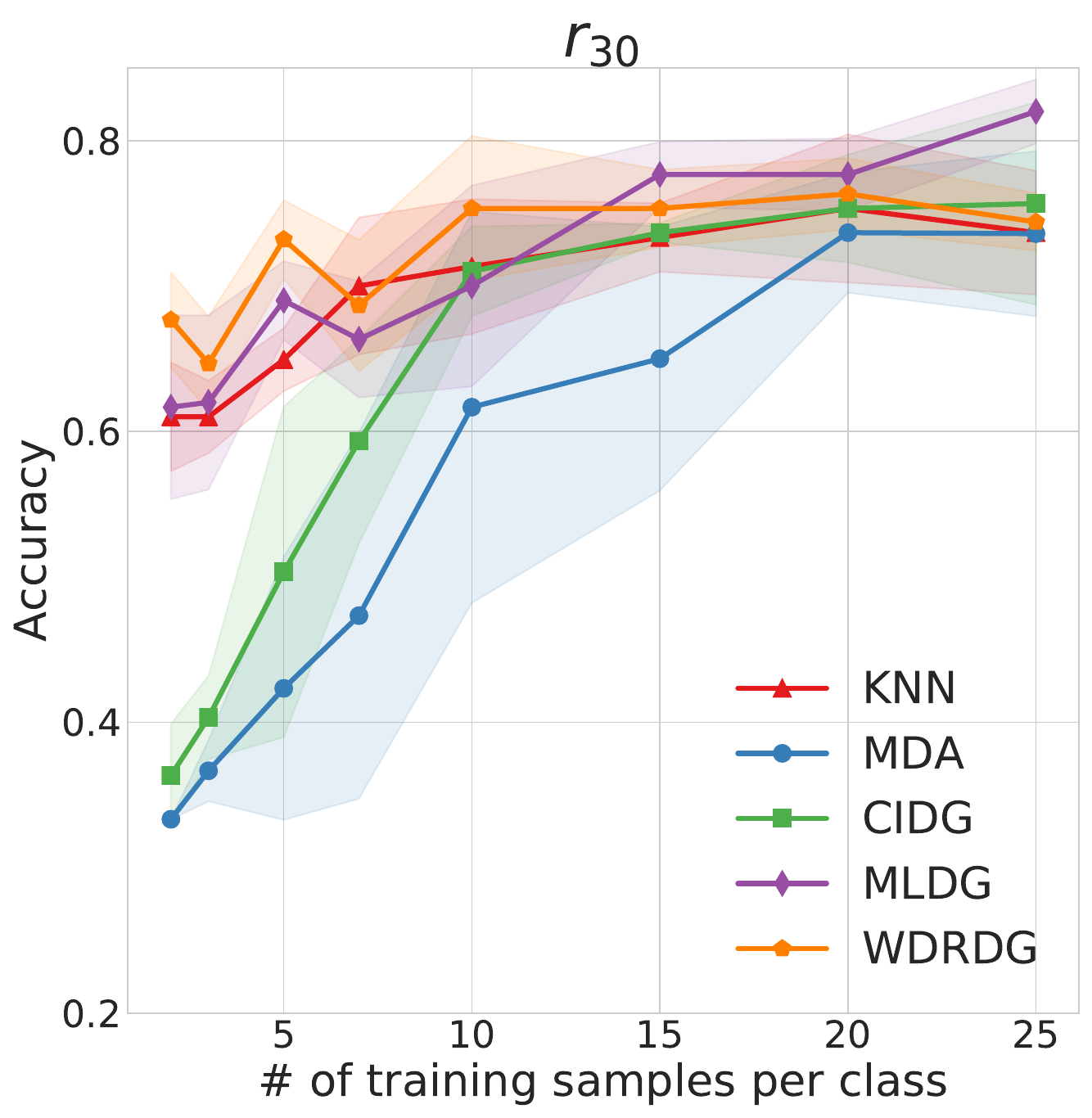}\label{fig:rmnist_results_30}}
  {\includegraphics[width=0.23\linewidth]{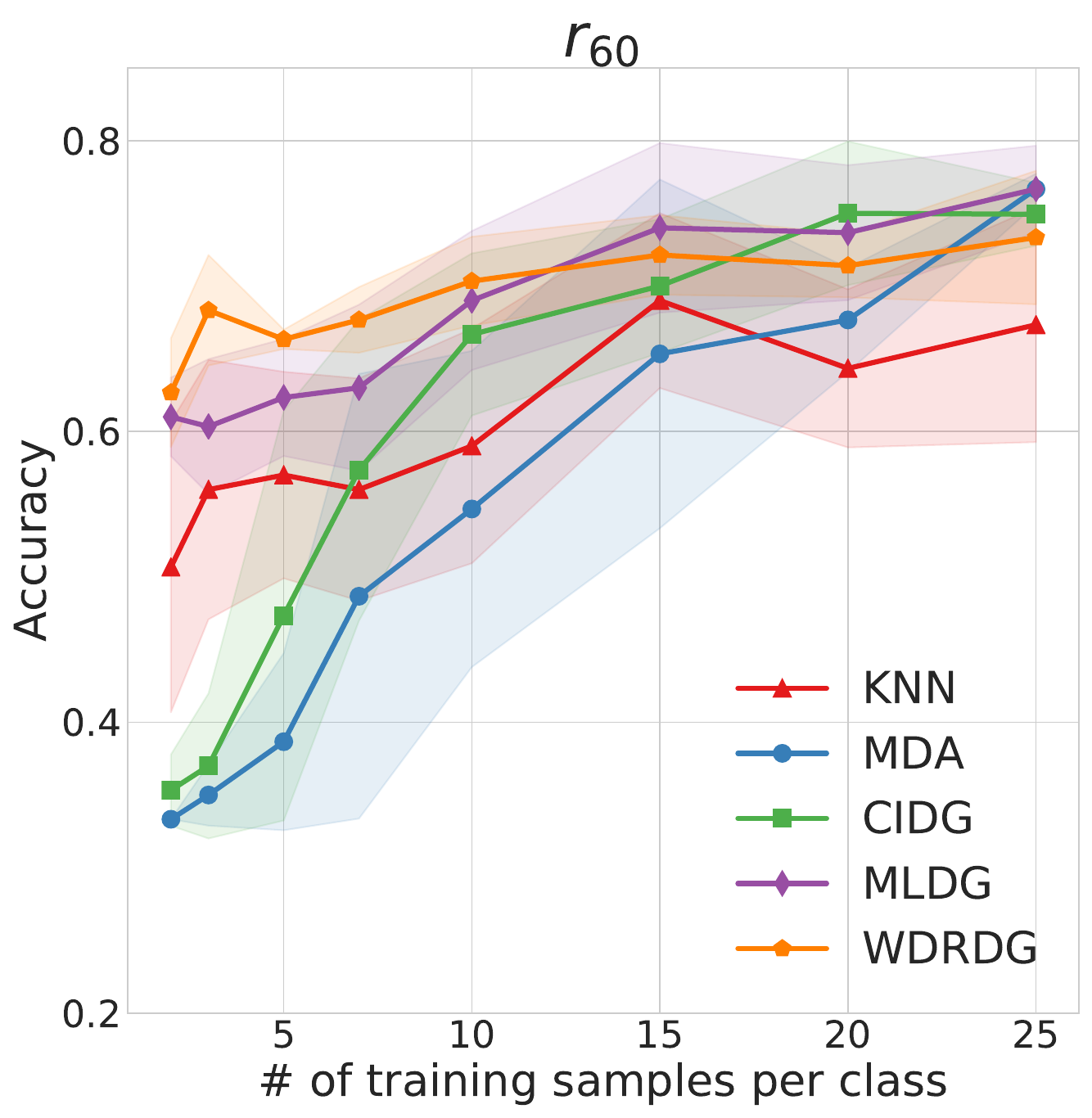}\label{fig:rmnist_results_60}}
  {\includegraphics[width=0.23\linewidth]{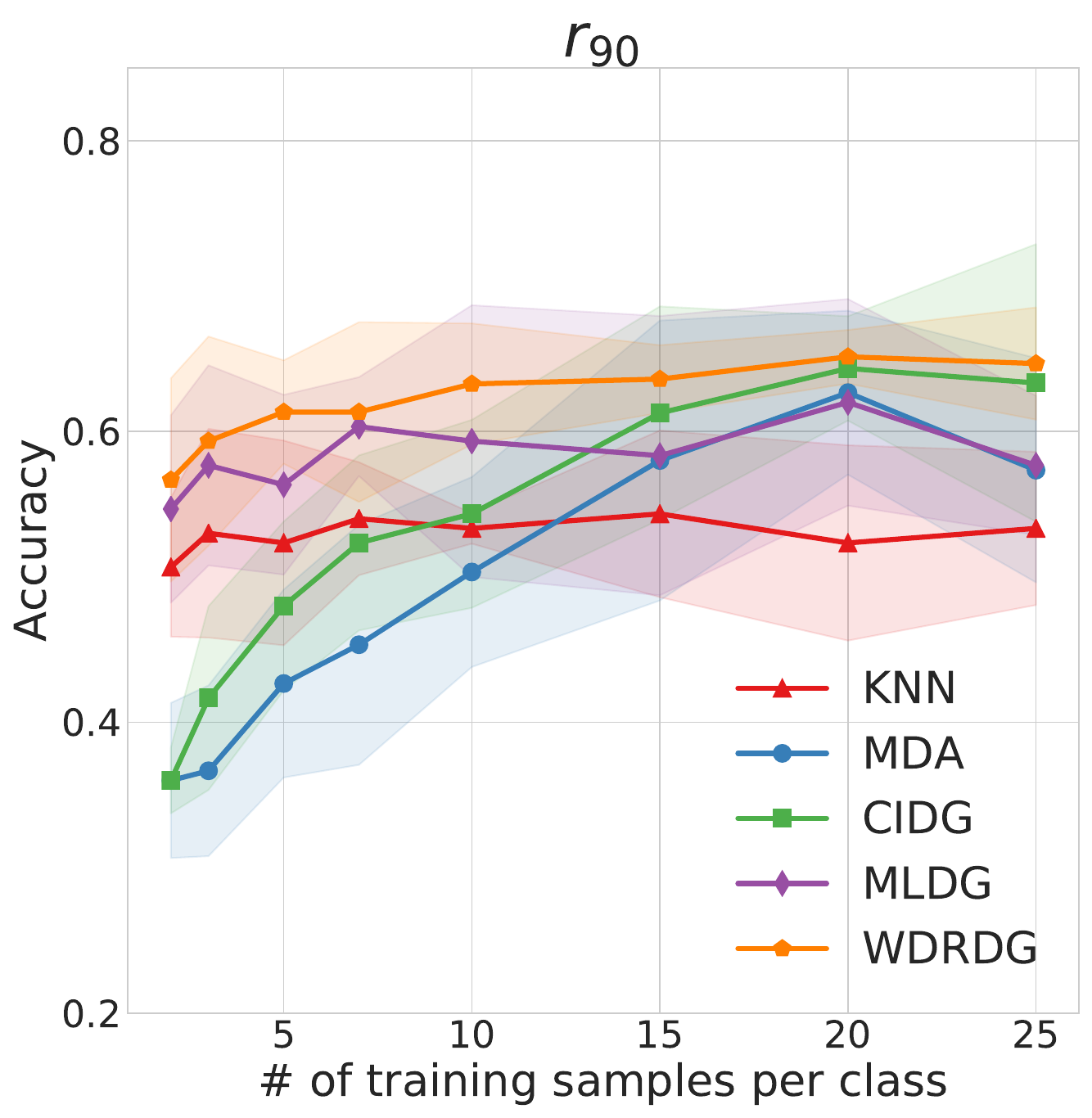}\label{fig:rmnist_results_90}}
\caption{Performance comparison for the VLCS, PACS and Rotated MNIST dataset under different size of training samples per class. Each row shows the results for a dataset, and each column shows the generalization result for a certain target domain. Average performance of five methods are represented by different colors, and the corresponding shadow shows the standard deviation of 5 experimental trials. 
Our WDRDG framework outperforms KNN, MDA and CIDG with higher accuracy and smaller standard deviation.
Also, it has more advantage over MLDG especially when the source training sample size is limited. For example, WDRDG outperforms MLDG by up to 
$46.79\%$ when the target domain is Caltech-101 in the VLCS dataset, by up to $20.95\%$ for target domain Art Painting in the PACS dataset, 
and by up to  $20.71\%$ for target domain $r_{0}$ in the Rotated MNIST dataset  with training sample size of 2 for each class.
}
\label{fig:all_results}
\end{figure*}

\begin{figure*}
    \centering
  \subfloat[VLCS]{\includegraphics[width=0.26\linewidth]{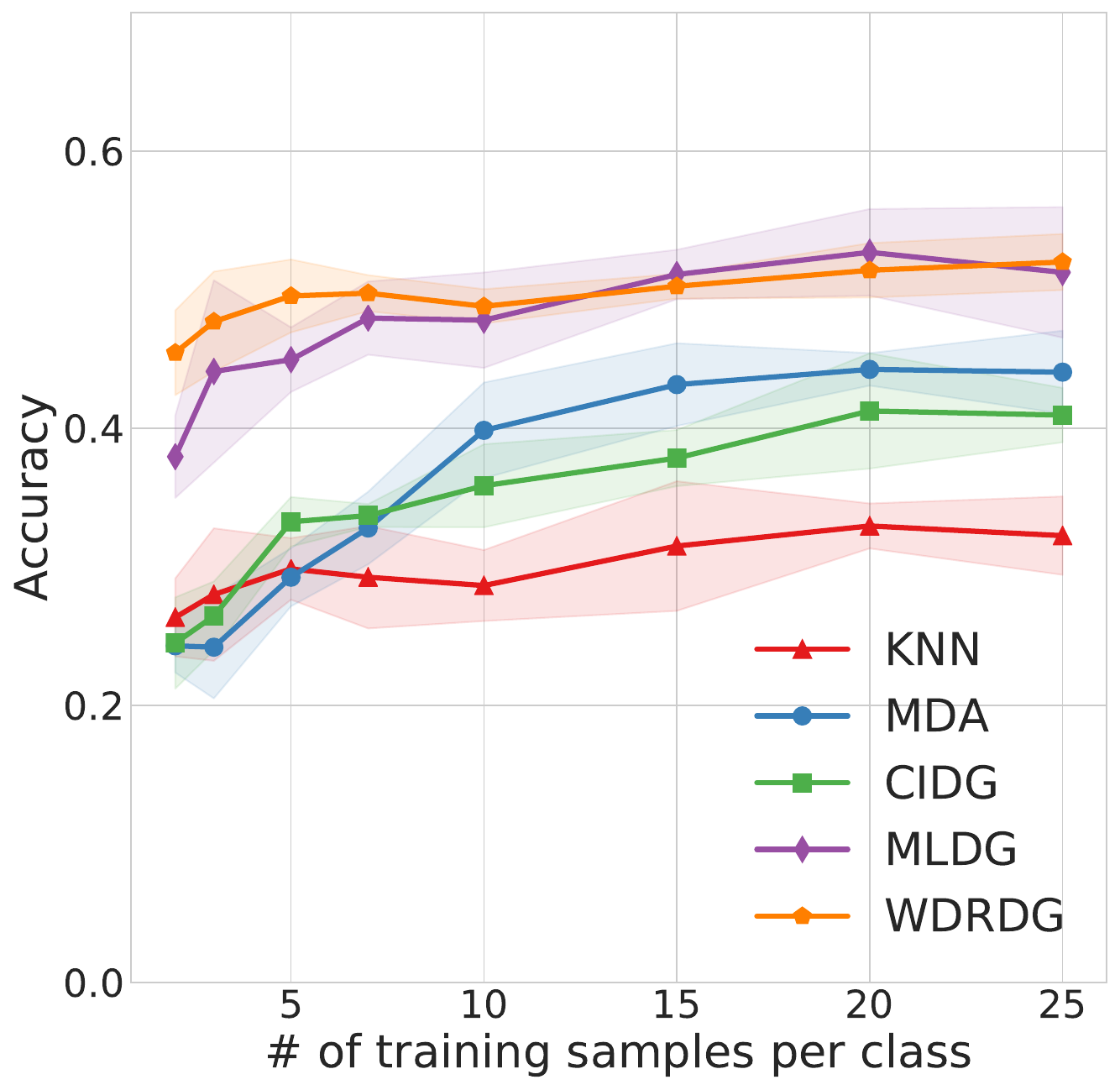}\label{subfig:vlcs_ave}}
  \subfloat[PACS]{\includegraphics[width=0.26\linewidth]{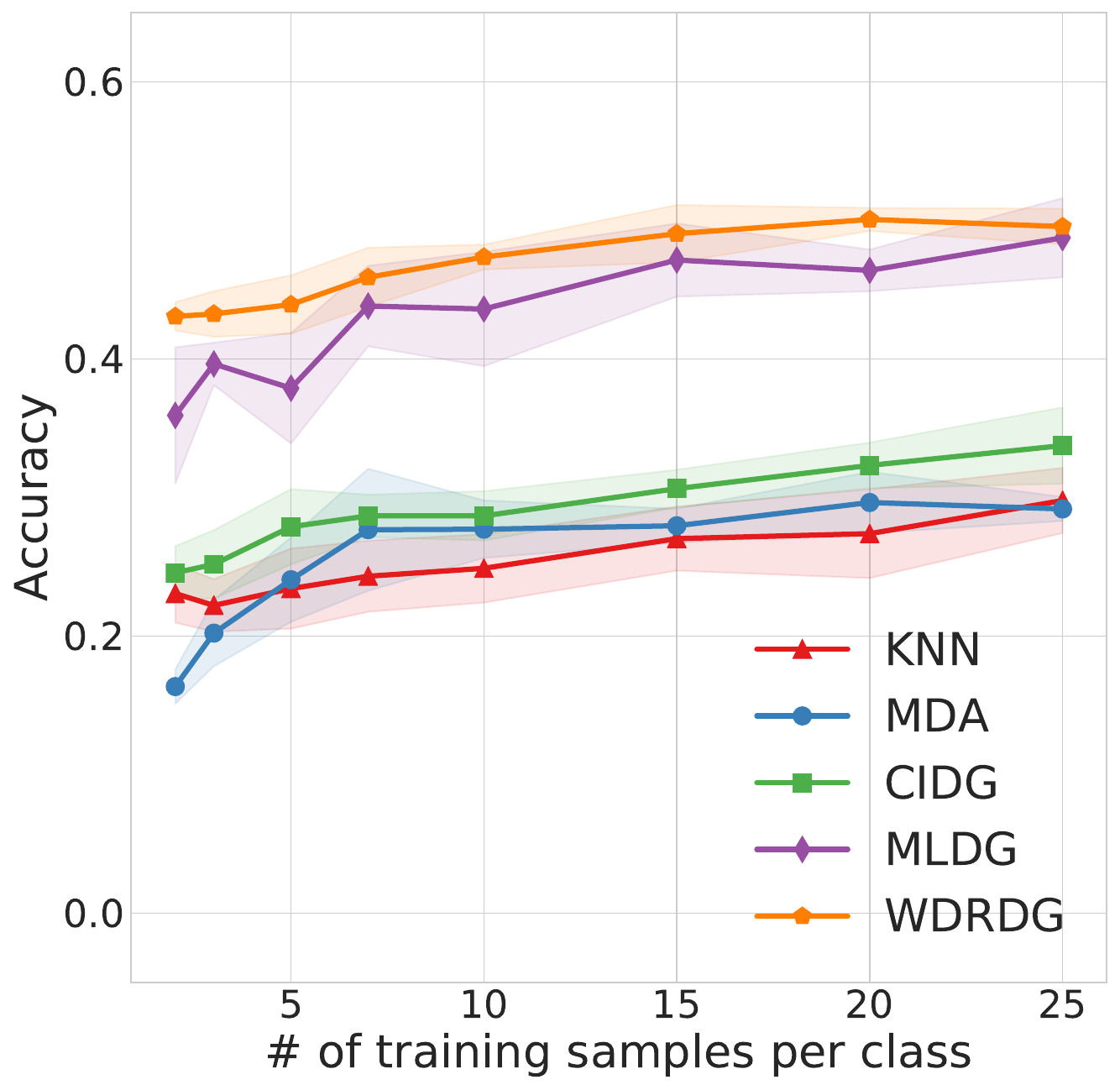}\label{subfig:pacs_ave}}
  \subfloat[Rotated MNIST]{\includegraphics[width=0.26\linewidth]{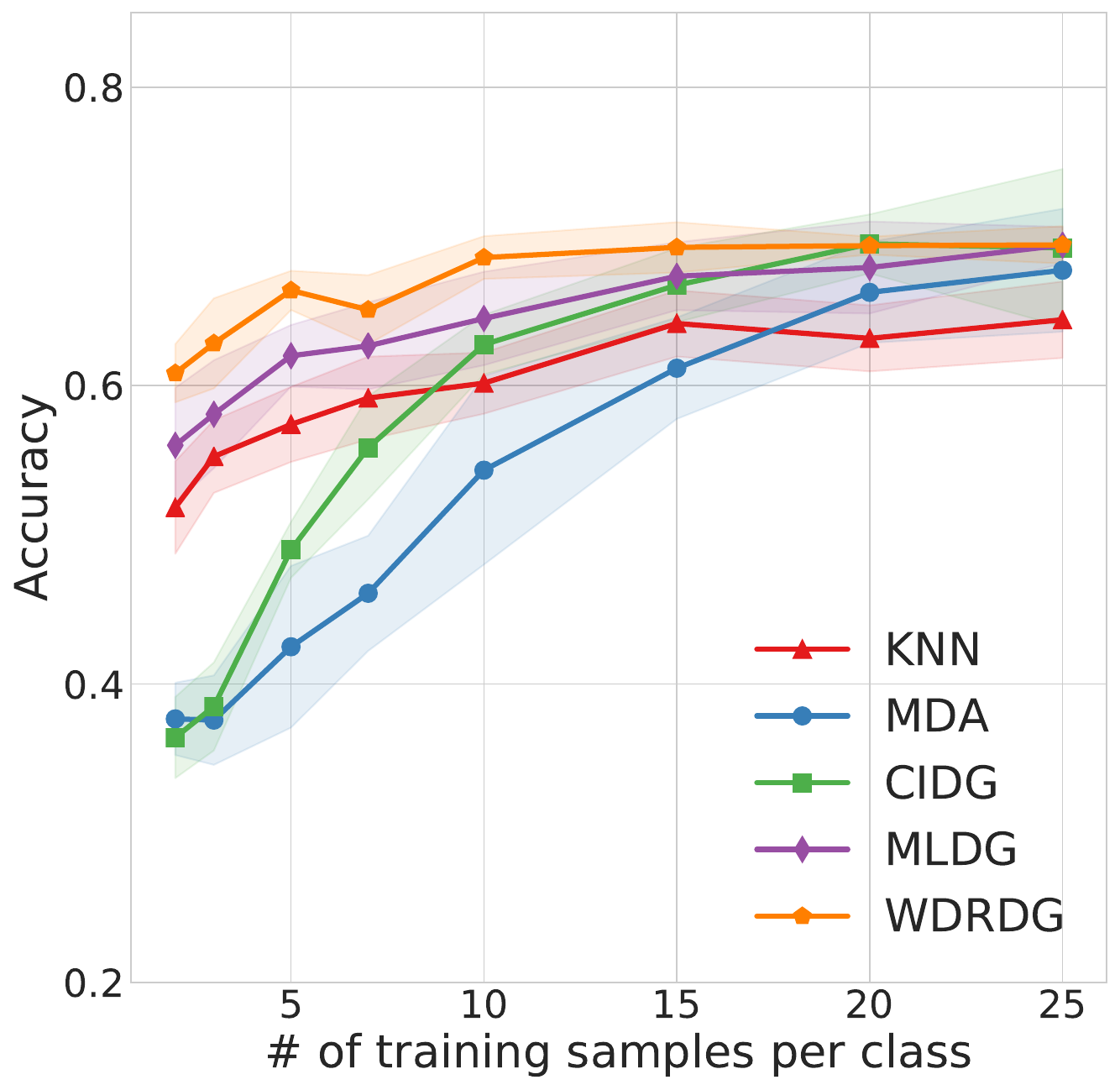}\label{subfig:rmnist_ave}}
    \caption{Average generalization performance of different methods on the VLCS, PACS and Rotated MNIST dataset. As the training sample size increases, all methods obtain better performance. Our WDRDG framework outperforms other baselines, especially in few-shot settings. When the sample size is less than 10 per class, WDRDG provides at least $3.75\%$, $4.73\%$, $3.86\%$ better generalization ability than others on the VLCS, PACS and Rotated MNIST dataset, respectively.}
    \label{fig:results_ave}
\end{figure*}

In this section, we present the results for domain generalization on all three datasets.
When each domain serves as the target domain, the results are shown in Figure \ref{fig:all_results}, with the plotted lines representing the average performance over $5$ trials and the shaded area representing the corresponding standard deviation.

For the VLCS dataset, we report the results in the first row in Figure \ref{fig:all_results}.
In all four cases when each domain serves as the unseen target domain, our method achieves better classification accuracy and standard deviation than other methods when the training sample size for each class is very few, i.e., 2, 3 or 5. 
The advantage over MLDG then levels off as the sample size reaches to over 10 per class.
The performance improvement against MLDG reaches as high as $6.53\%$, $11.89\%$, $46.79\%$, $22.54\%$ with only 2 training samples for each class when the target domain is PASCAL VOC2007, LabelMe, Caltech-101 and SUN09, respectively, which confirms that our method is efficient for few-shot cases.

The second row of Figure \ref{fig:all_results} reports the classification accuracy results for the PACS dataset.
The proposed WDRDG achieves the best results in accuracy and standard deviation when the target domain is Art Painting, Cartoon, or Sketch using different training sample size, and MLDG outperforms WDRDG when the target domain is Photos with the sample size 15 for each class. 
WDRDG outperforms MLDG by up to $19.81\%$, $20.95\%$, $18.68\%$, $20.35\%$ for each target domain when the training sample size is 2. 
This validates the effect of our method when the training sample size is limited. 
The improvement of WDRDG over other methods on the PACS dataset is relatively larger compared with the improvements on the VLCS dataset. This improvement is especially obvious over MDA and CIDG when the target domain is Sketch, shown in the fourth column of the second row in Figure \ref{fig:all_results}.
This may because that the differences among domains are greater in PACS where the image styles are obviously different compared with in VLCS, where samples from different domains are real-world images collected from different perspectives or scales. This demonstrates that our WDRDG could better handle scenarios with larger unseen domain shift.

The results for the Rotated MNIST dataset in the third row of  Figure \ref{fig:all_results} also yield similar conclusions. As the training sample size increases, almost all methods converges to the same accuracy for different target domain.
When the training sample size is smaller, i.e., the training sample per class for each source domain is $2,3,5,7$, the advantage of  our proposed framework is more obvious. 
WDRDG outperforms MLDG by $20.71\%$, $9.73\%$, $2.73\%$, $3.66\%$ when the training sample size is 2 for each class for target domain $r_{0}$, $r_{30}$, $r_{60}$, and $r_{90}$, respectively.
When the training sample size is big, e.g., the training sample per class for each source domain is $25$, even simple KNN method performs well. 
This is consistent with the analysis in the above two datasets.

Figure \ref{fig:results_ave} reports the average performance of different target domains on the three datasets.
Overall, our method is the most stable under different numbers of training samples, with narrower shadow band of standard deviation.
As the size of training samples gets bigger, all methods have the tendency of performing better.
On the PACS and Rotated MNIST dataset, WDRDG achieves the best average performance under different training sample size compared with other methods.
On the VLCS dataset, WDRDG also achieves pretty good accuracies with smaller standard deviation.
In addition, our method shows more advantage over others in few-shot settings. When given training samples are limited to less than 10 (i.e., 2, 3, 5, 7 in our experiments) per class, WDRDG provides at least $3.75\%$, $4.73\%$, $3.86\%$ better generalization ability than others on the VLCS, PACS and Rotated MNIST dataset, respectively.



\begin{table*}
\caption{The effect of the optimal transport-based test-time adaptation (TTA) for adaptive inference on the VLCS dataset. WDRDG with the TTA module results in better performance when using a different number of training samples.
}
\label{table:ablation_VLCS}
\centering
\begin{tabular}{cc ccccc}
\hline
\multirow{2}*{training sample size/class} & \multirow{2}*{Method} &\multicolumn{5}{c}{Target} \\
\cline{3-7}
& & V &  L & C  & S & Average \\
\hline
\multirow{2}*{5} & WDRDG (w/o. TTA) &0.516	&0.372	&0.554	&0.356 & 0.450
\\
& WDRDG (w. TTA)   &0.582	&0.448	&0.494	&0.458	&\textbf{0.496}\\
\hline
\multirow{2}*{10} & WDRDG (w/o. TTA) &0.540 & 0.402 & 0.516 & 0.334 & 0.448\\
& WDRDG (w. TTA)   &0.546 &0.410 &0.546 &0.450 &\textbf{0.488}\\
\hline
\multirow{2}*{15} & WDRDG (w/o. TTA) &0.510	&0.378	&0.67	&0.39 & 0.487
\\
& WDRDG (w. TTA)   &0.568	&0.438	&0.564	&0.440	&\textbf{0.503}
\\
\hline
\end{tabular} 

\end{table*}

\begin{table*}
\caption{The effect of the optimal transport-based test-time adaptation (TTA) for adaptive inference on the PACS dataset. WDRDG with the TTA module results in better performance when using a different number of training samples.}
\label{table:ablation_PACS}
\centering
\begin{tabular}{cc ccccc}
\hline
\multirow{2}*{training sample size/class} & \multirow{2}*{Method} &\multicolumn{5}{c}{Target} \\
\cline{3-7}
& & P &  A & C  & S & Average \\
\hline
\multirow{2}*{5} & WDRDG (w/o. TTA) &0.504	&0.350	&0.471	&0.237	&0.391\\
& WDRDG (w. TTA)   &0.514	&0.403	&0.441	&0.399	&\textbf{0.439}
\\
\hline
\multirow{2}*{10} & WDRDG (w/o. TTA) &0.559 &0.374 &0.480 &0.259 &0.418\\
& WDRDG (w. TTA)   &0.556 &0.421 &0.519 &0.409 &\textbf{0.476}\\
\hline
\multirow{2}*{15} & WDRDG (w/o. TTA) &0.549	&0.404	&0.491	&0.251	&0.424\\
& WDRDG (w. TTA)   &0.533	&0.477	&0.475	&0.462	&\textbf{0.487}
\\
\hline
\end{tabular} 
\end{table*}

\subsection{Ablation Study for the Test-time Adaptation}
To explore the effectiveness of the test-time adaptation based on optimal transport, we compare our framework with and without this adaptive inference module. 
For the non-adaptive inference, the nearest neighbor for any test sample from the target domain is found by the simple 1-NN over barycenter samples.
We compare the results of using training sample size of $5,10,15$ per class for each source domain. 

From the results in Table \ref{table:ablation_VLCS},  \ref{table:ablation_PACS}, and \ref{table:ablation_rMNIST} for VLCS, PACS and Rotated MNIST dataset, respectively, 
we can make several observations.
Our WDRDG framework with the adaptive inference module results in better average performance for all three datasets, with up to $10.22\%$ higher mean accuracy for the VLCS dataset with 5 training samples per class, $14.86\%$ performance improvement for the PACS dataset with 15 training samples per class, and $13.98\%$ improvements for the Rotated MNIST dataset with 15 training samples per class.
Note that when the target domain is Sketch on the PACS dataset, the improvements are especially obvious compared with other targets, reaching $68.35\%$, $57.92\%$, and $84.06\%$ when the training sample size for each class is $5,10,15$, respectively.
Similar results could be found on the Rotated MNIST dataset when the target domain is $r_{0}$ or $r_{90}$ when the training sample size per class is $10$ or $15$, with up to $19.32\%$ performance improvements. This improvement is more obvious compared with other targets $r_{30}$ or $r_{60}$, which obtains up to $15.31\%$ performance improvements using the adaptive inference module.
One thing they share in common is these target domains are more different with given source domains, which shows larger unseen distribution shifts. 
This validates the robustness of our adaptive inference module for even harder, unseen target domains.

\subsection{Analysis of Imbalanced Classes among Source Domains}
\begin{table*}
\caption{The effect of the optimal transport-based test-time adaptation (TTA) for adaptive inference on the Rotated MNIST dataset. WDRDG with the TTA module results in better performance when using a different number of training samples.}
\label{table:ablation_rMNIST}
\centering
\begin{tabular}{cc ccccc}
\hline
\multirow{2}*{training sample size/class} & \multirow{2}*{Method} &\multicolumn{5}{c}{Target} \\
\cline{3-7}
& & $r_{0}$ &  $r_{30}$ & $r_{60}$  & $r_{90}$ & Average \\
\hline
\multirow{2}*{5} & WDRDG (w/o. TTA) &0.593	&0.640	&0.577	&0.553	&0.591
\\
& WDRDG (w. TTA)   &0.647	&0.732	&0.663	&0.613	&\textbf{0.664}
\\
\hline
\multirow{2}*{10} & WDRDG (w/o. TTA) &0.567 &0.690 &0.647 &0.557 &0.615\\
& WDRDG (w. TTA)   &0.654 &0.753 &0.703 &0.633 &\textbf{0.686}\\
\hline
\multirow{2}*{15} & WDRDG (w/o. TTA) &0.567	&0.653	&0.677 &0.533	&0.608
\\
& WDRDG (w. TTA)   &0.660	&0.753	&0.721	&0.636	&\textbf{0.693}
\\
\hline
\end{tabular} 
\end{table*}
\begin{figure*}
  \centering
    {\includegraphics[width=0.2\linewidth]{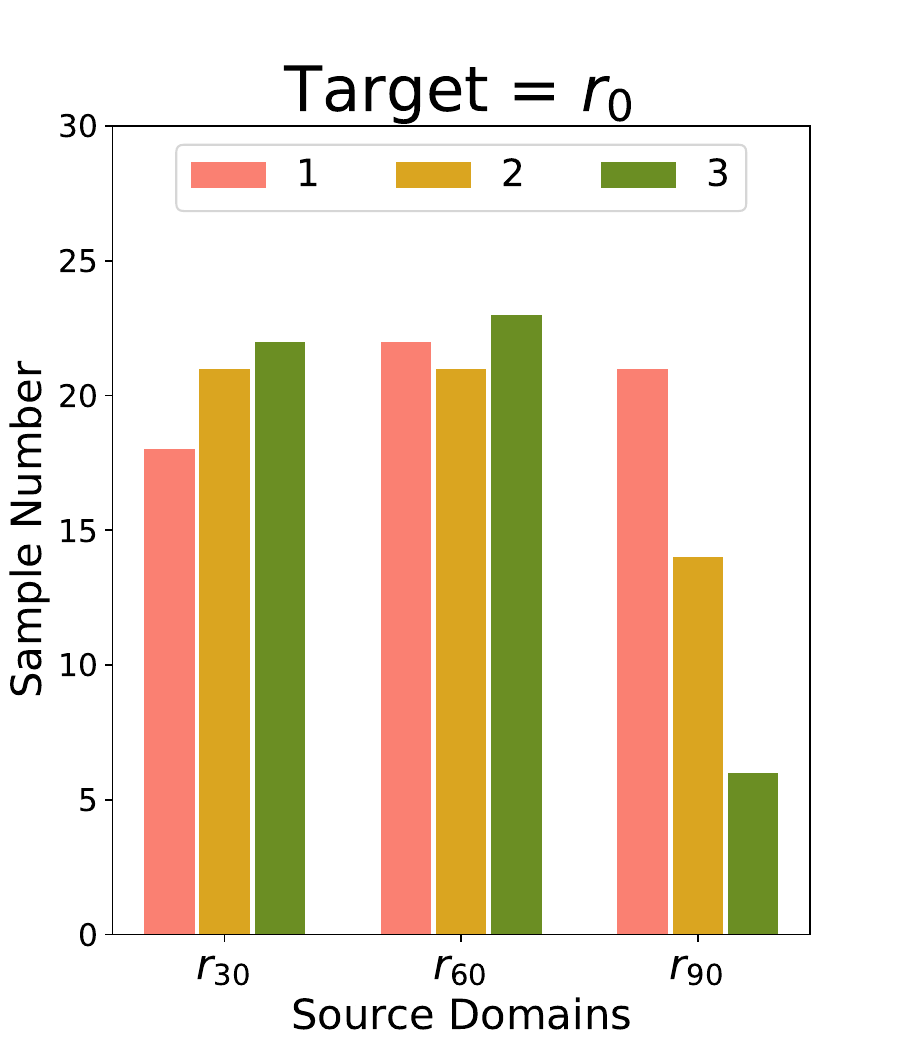}\label{subfig:rmnist_results_0}}
    {\includegraphics[width=0.2\linewidth]{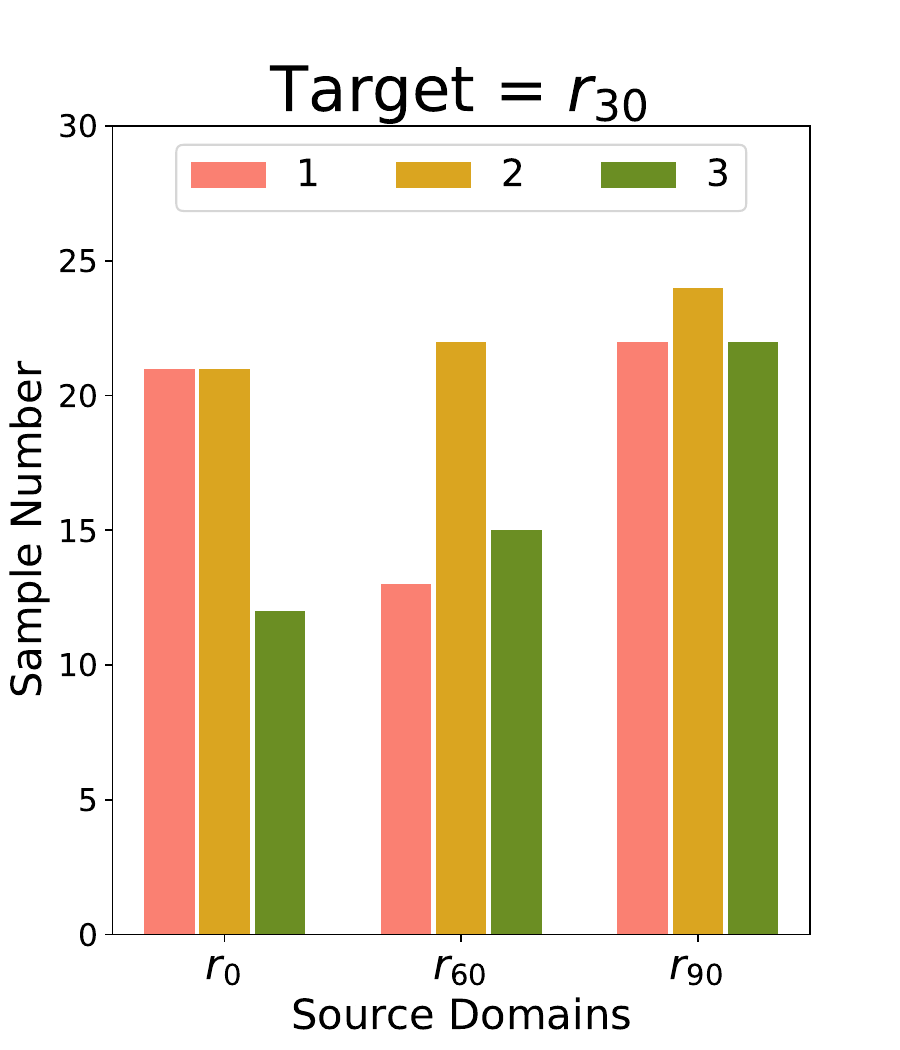}\label{subfig:rmnist_results_30}}
    {\includegraphics[width=0.2\linewidth]{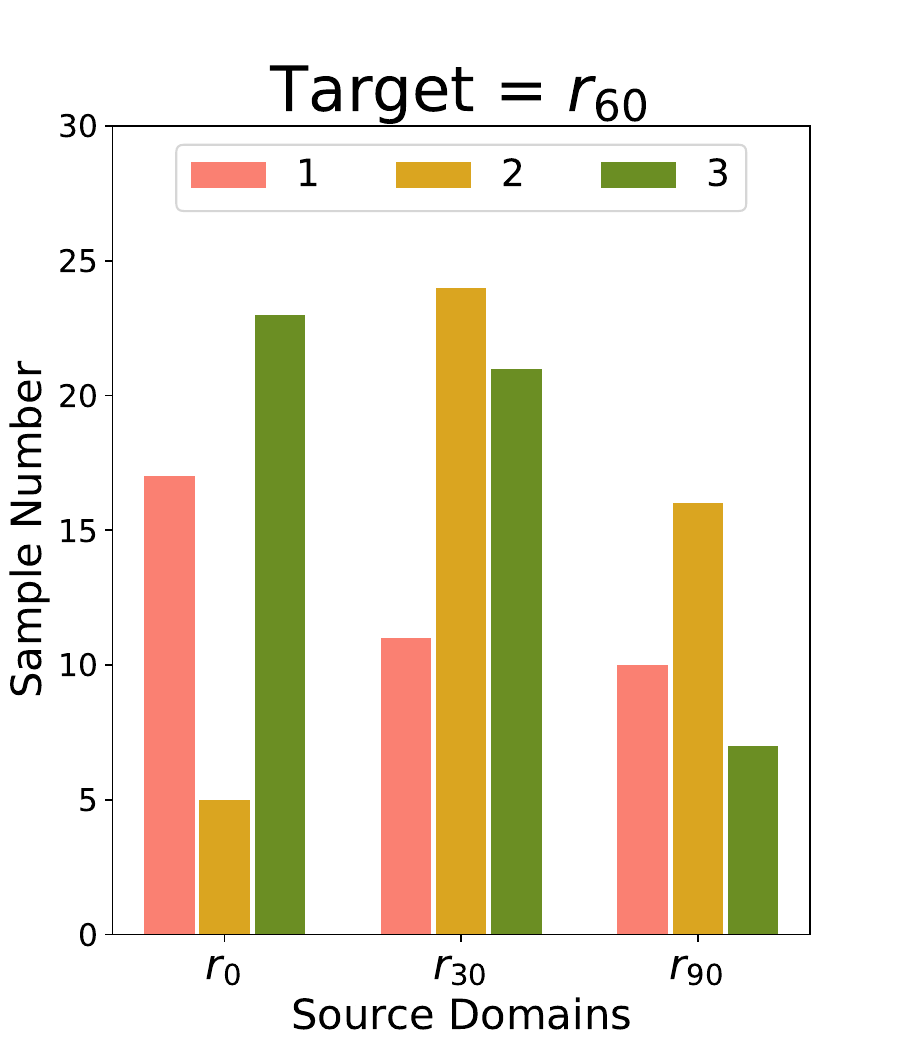} \label{subfig:rmnist_results_60}}
    {\includegraphics[width=0.2\linewidth]{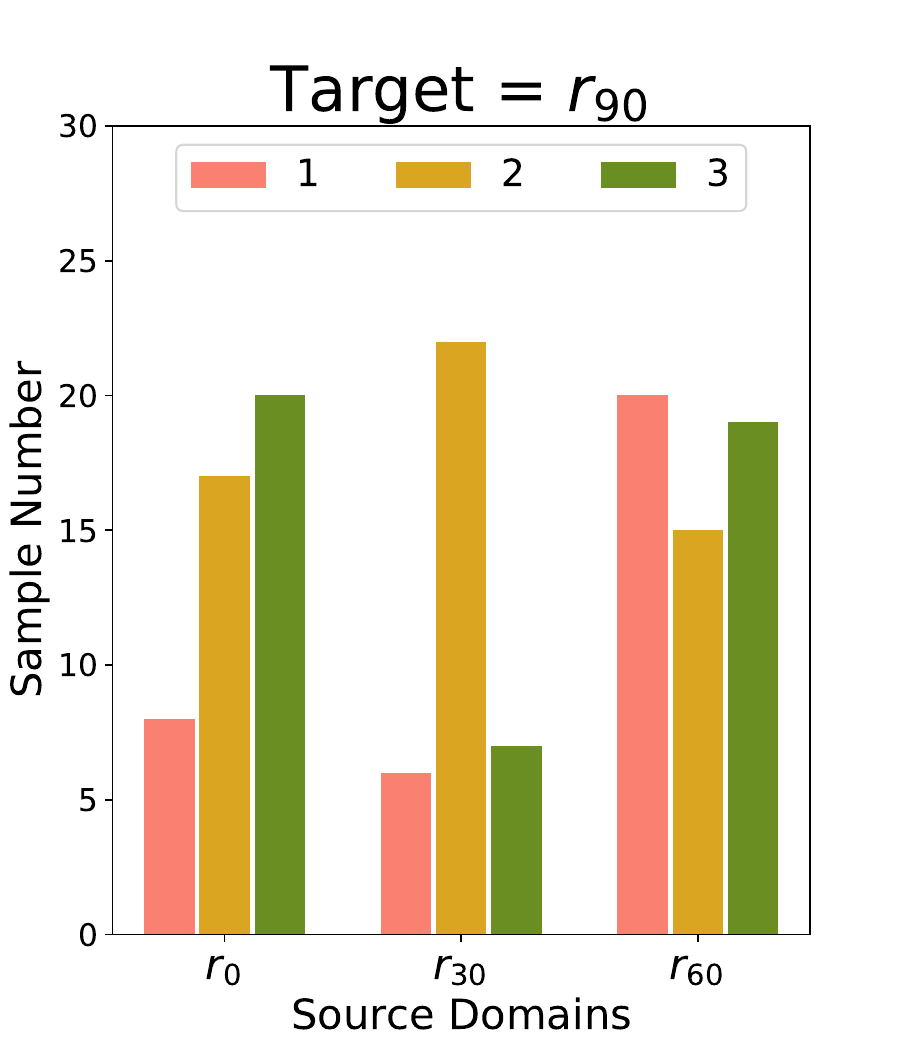}\label{subfig:rmnist_results_90}}
  \caption{Visualization of random sample size for each class in source domains when a different domain serves as the target domain in the Rotated MNIST dataset. For each source domain, the number of samples for different classes are shown in different colors. There are cases when different classes have similar sample number, e.g., Class 1 and 2 of source domain $r_0$ when target domain is $r_{30}$, and also cases when different classes have quite different number of samples, e.g., in source domain $r_{90}$ when target domain is $r_{0}$.}
  \label{fig:rmnist_imbalance_distribution}
\end{figure*}

\begin{figure}
    \centering
    \includegraphics[scale=0.3]{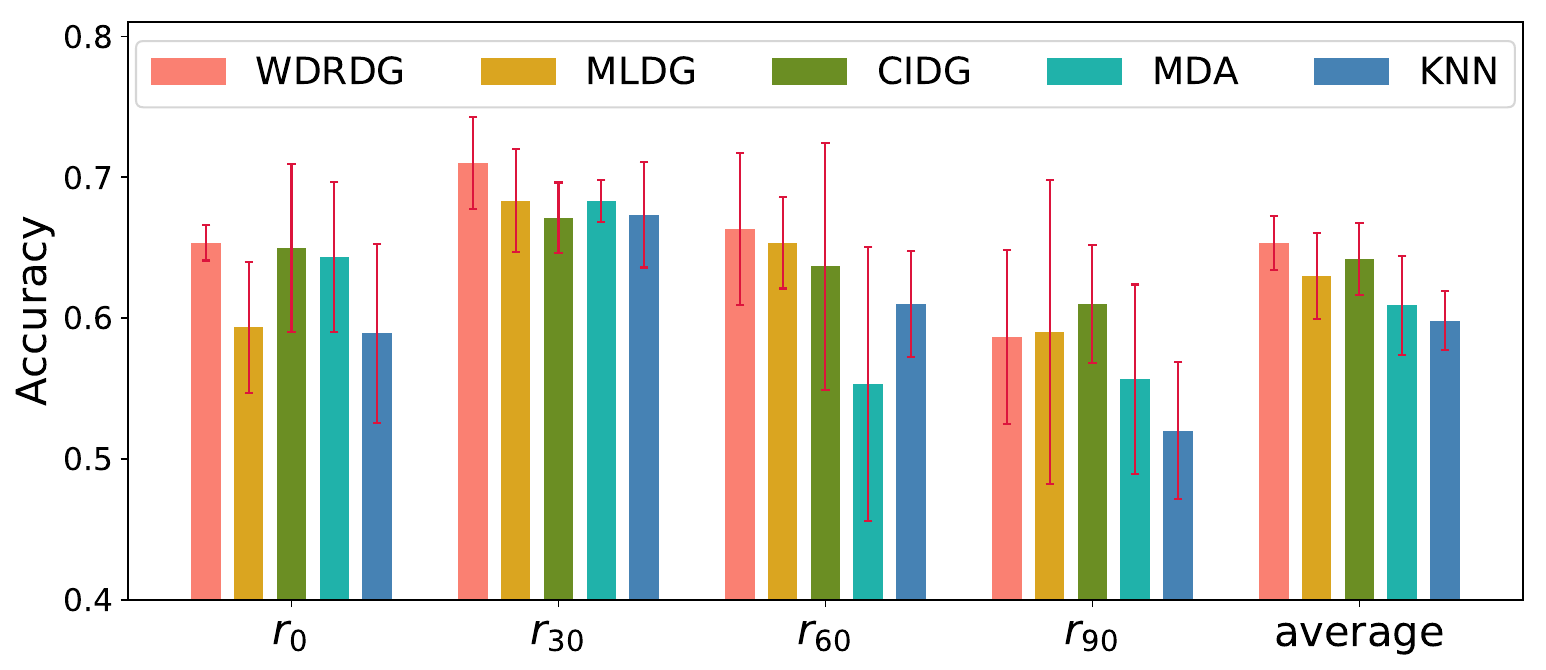}
    \caption{The performance of WDRDG and four compared methods on the Rotated MNIST dataset with different class prior distributions across source domains.  WDRDG outperforms other baselines by at least
$0.51\%$, $3.90\%$, $1.53\%$ when the target domain is $r_{0}$, $r_{30}$, $r_{60}$, respectively, and achieves similar accuracies with MLDG but with smaller deviation when the target domain is $r_{90}$.}
    \label{fig:rmnist_imbalance_results}
\end{figure}
In previous experiments, we actually assume the training sample size per class in the source domains are the same under the setting of no class prior distribution shift, i.e., the distribution of $P(Y)$ is the same across all source domains. 
To show the feasibility of extending our framework to scenarios with class prior distribution shift, we further conduct experiments when the categories in source domains are imbalanced, i.e., there are shifts among $P(Y)$ of different domains. 

We randomly sample the training sample size for each class from $[5,25)$ on the Rotated MNIST dataset here. The distribution of sample number for each class when each domain is chosen as the target domain is shown in Figure \ref{fig:rmnist_imbalance_distribution}.
There are cases when different classes have similar sample number, e.g., in source domain $r_{90}$ when the target domain is $r_{30}$, or in source domain $r_{60}$ when the target domain is $r_{0}$.  
In other source domains, different classes may have quite different number of samples, e.g., in source domain $r_{90}$ when target domain is $r_{0}$, or in source domain $r_{0}$ when target domain is $r_{60}$.
We compare our framework WDRDG with other methods, and the results are shown in Figure \ref{fig:rmnist_imbalance_results}.
When the target domain is $r_{90}$, our method achieves similar accuracies with MLDG but with smaller deviation, while in other cases WDRDG outperforms other baselines by at least
$0.51\%$, $3.90\%$, $1.53\%$ when the target domain is $r_{0}$, $r_{30}$, $r_{60}$, respectively. 
Our framework outperforms other methods on average with smaller standard deviation, which validates the generalization ability of our framework when the source domains have class prior distribution shift.

\section{Conclusion}
In this research, we proposed a novel framework for domain generalization to enhance model robustness when labeled training data of source domains are limited.
We formulated the distributional shifts for each class with class-specific Wasserstein uncertainty sets and optimized the model over the worst-case distributions residing in the uncertainty sets via distributionally robust optimization. 
To reduce the difference between source and target domains, we proposed a test-time domain adaptation module through optimal transport to make adaptive inference for unseen target data. We found that our domain generalization framework with this adaptive inference module works better when target domains are more different compared with source domains. 
Experimental results on Rotated MNIST, PACS and VLCS datasets demonstrate that our proposed WDRDG framework could learn a robust model for unseen target domains based on limited source data, and we also showed that its advantage is more obvious in few-shot settings. 
To perfect this work in the future,  we would study the usage of class priors in constructing more realistic uncertainty sets, and explore measurable relationship among source domains to better leverage the source distributions to model possible target distributions.

\ifCLASSOPTIONcompsoc
  \section*{Acknowledgments}
\else
  \section*{Acknowledgment}
  This study is supported in part by the Tsinghua SIGS Scientific Research Start-up Fund (Grant QD2021012C) and Natural Science Foundation of China (Grant 62001266).
\fi


\ifCLASSOPTIONcaptionsoff
  \newpage
\fi


\bibliographystyle{IEEEtran}
\bibliography{references}

\begin{thebibliography}{10}
\providecommand{\url}[1]{#1}
\csname url@samestyle\endcsname
\providecommand{\newblock}{\relax}
\providecommand{\bibinfo}[2]{#2}
\providecommand{\BIBentrySTDinterwordspacing}{\spaceskip=0pt\relax}
\providecommand{\BIBentryALTinterwordstretchfactor}{4}
\providecommand{\BIBentryALTinterwordspacing}{\spaceskip=\fontdimen2\font plus
\BIBentryALTinterwordstretchfactor\fontdimen3\font minus \fontdimen4\font\relax}
\providecommand{\BIBforeignlanguage}[2]{{%
\expandafter\ifx\csname l@#1\endcsname\relax
\typeout{** WARNING: IEEEtran.bst: No hyphenation pattern has been}%
\typeout{** loaded for the language `#1'. Using the pattern for}%
\typeout{** the default language instead.}%
\else
\language=\csname l@#1\endcsname
\fi
#2}}
\providecommand{\BIBdecl}{\relax}
\BIBdecl

\bibitem{blanchard2011generalizing}
G.~Blanchard, G.~Lee, and C.~Scott, ``Generalizing from several related classification tasks to a new unlabeled sample,'' \emph{Advances in neural information processing systems}, vol.~24, pp. 2178--2186, 2011.

\bibitem{gong2019dlow}
R.~Gong, W.~Li, Y.~Chen, and L.~V. Gool, ``Dlow: Domain flow for adaptation and generalization,'' in \emph{Proceedings of the IEEE/CVF Conference on Computer Vision and Pattern Recognition}, 2019, pp. 2477--2486.

\bibitem{shi2020towards}
Y.~Shi, X.~Yu, K.~Sohn, M.~Chandraker, and A.~K. Jain, ``Towards universal representation learning for deep face recognition,'' in \emph{Proceedings of the IEEE/CVF Conference on Computer Vision and Pattern Recognition}, 2020, pp. 6817--6826.

\bibitem{zhou2020learning}
K.~Zhou, Y.~Yang, T.~Hospedales, and T.~Xiang, ``Learning to generate novel domains for domain generalization,'' in \emph{European conference on computer vision}.\hskip 1em plus 0.5em minus 0.4em\relax Springer, 2020, pp. 561--578.

\bibitem{zhou2021learning}
K.~Zhou, Y.~Yang, A.~Cavallaro, and T.~Xiang, ``Learning generalisable omni-scale representations for person re-identification,'' \emph{IEEE Transactions on Pattern Analysis and Machine Intelligence}, 2021.

\bibitem{yue2019domain}
X.~Yue, Y.~Zhang, S.~Zhao, A.~Sangiovanni-Vincentelli, K.~Keutzer, and B.~Gong, ``Domain randomization and pyramid consistency: Simulation-to-real generalization without accessing target domain data,'' in \emph{Proceedings of the IEEE/CVF International Conference on Computer Vision}, 2019, pp. 2100--2110.

\bibitem{volpi2018generalizing}
R.~Volpi, H.~Namkoong, O.~Sener, J.~C. Duchi, V.~Murino, and S.~Savarese, ``Generalizing to unseen domains via adversarial data augmentation,'' in \emph{NeurIPS}, 2018.

\bibitem{shao2019multi}
R.~Shao, X.~Lan, J.~Li, and P.~C. Yuen, ``Multi-adversarial discriminative deep domain generalization for face presentation attack detection,'' in \emph{Proceedings of the IEEE/CVF Conference on Computer Vision and Pattern Recognition}, 2019, pp. 10\,023--10\,031.

\bibitem{balaji2018metareg}
Y.~Balaji, S.~Sankaranarayanan, and R.~Chellappa, ``Metareg: Towards domain generalization using meta-regularization,'' \emph{Advances in Neural Information Processing Systems}, vol.~31, pp. 998--1008, 2018.

\bibitem{marzinotto2019robust}
G.~Marzinotto, G.~Damnati, F.~B{\'e}chet, and B.~Favre, ``Robust semantic parsing with adversarial learning for domain generalization,'' \emph{arXiv preprint arXiv:1910.06700}, 2019.

\bibitem{stepanov2014towards}
E.~Stepanov and G.~Riccardi, ``Towards cross-domain pdtb-style discourse parsing,'' in \emph{Proceedings of the 5th International Workshop on Health Text Mining and Information Analysis (Louhi)}, 2014, pp. 30--37.

\bibitem{fried2019cross}
D.~Fried, N.~Kitaev, and D.~Klein, ``Cross-domain generalization of neural constituency parsers,'' \emph{arXiv preprint arXiv:1907.04347}, 2019.

\bibitem{li2018learning}
D.~Li, Y.~Yang, Y.-Z. Song, and T.~M. Hospedales, ``Learning to generalize: Meta-learning for domain generalization,'' in \emph{Thirty-Second AAAI Conference on Artificial Intelligence}, 2018.

\bibitem{muandet2013domain}
K.~Muandet, D.~Balduzzi, and B.~Sch{\"o}lkopf, ``Domain generalization via invariant feature representation,'' in \emph{International Conference on Machine Learning}.\hskip 1em plus 0.5em minus 0.4em\relax PMLR, 2013, pp. 10--18.

\bibitem{ghifary2016scatter}
M.~Ghifary, D.~Balduzzi, W.~B. Kleijn, and M.~Zhang, ``Scatter component analysis: A unified framework for domain adaptation and domain generalization,'' \emph{IEEE transactions on pattern analysis and machine intelligence}, vol.~39, no.~7, pp. 1414--1430, 2016.

\bibitem{blanchard2017domain}
G.~Blanchard, A.~A. Deshmukh, U.~Dogan, G.~Lee, and C.~Scott, ``Domain generalization by marginal transfer learning,'' \emph{arXiv preprint arXiv:1711.07910}, 2017.

\bibitem{grubinger2015domain}
T.~Grubinger, A.~Birlutiu, H.~Sch{\"o}ner, T.~Natschl{\"a}ger, and T.~Heskes, ``Domain generalization based on transfer component analysis,'' in \emph{International Work-Conference on Artificial Neural Networks}.\hskip 1em plus 0.5em minus 0.4em\relax Springer, 2015, pp. 325--334.

\bibitem{li2018domain2}
Y.~Li, M.~Gong, X.~Tian, T.~Liu, and D.~Tao, ``Domain generalization via conditional invariant representations,'' in \emph{Proceedings of the AAAI Conference on Artificial Intelligence}, vol.~32, no.~1, 2018.

\bibitem{hu2020domain}
S.~Hu, K.~Zhang, Z.~Chen, and L.~Chan, ``Domain generalization via multidomain discriminant analysis,'' in \emph{Uncertainty in Artificial Intelligence}.\hskip 1em plus 0.5em minus 0.4em\relax PMLR, 2020, pp. 292--302.

\bibitem{zhou2020domain}
F.~Zhou, Z.~Jiang, C.~Shui, B.~Wang, and B.~Chaib-draa, ``Domain generalization with optimal transport and metric learning,'' \emph{arXiv preprint arXiv:2007.10573}, 2020.

\bibitem{peng2019moment}
X.~Peng, Q.~Bai, X.~Xia, Z.~Huang, K.~Saenko, and B.~Wang, ``Moment matching for multi-source domain adaptation,'' in \emph{Proceedings of the IEEE/CVF international conference on computer vision}, 2019, pp. 1406--1415.

\bibitem{motiian2017unified}
S.~Motiian, M.~Piccirilli, D.~A. Adjeroh, and G.~Doretto, ``Unified deep supervised domain adaptation and generalization,'' in \emph{Proceedings of the IEEE international conference on computer vision}, 2017, pp. 5715--5725.

\bibitem{li2018domain}
H.~Li, S.~J. Pan, S.~Wang, and A.~C. Kot, ``Domain generalization with adversarial feature learning,'' in \emph{Proceedings of the IEEE Conference on Computer Vision and Pattern Recognition}, 2018, pp. 5400--5409.

\bibitem{Gong_2019_CVPR}
R.~Gong, W.~Li, Y.~Chen, and L.~V. Gool, ``Dlow: Domain flow for adaptation and generalization,'' in \emph{Proceedings of the IEEE/CVF Conference on Computer Vision and Pattern Recognition (CVPR)}, June 2019.

\bibitem{li2018deep}
Y.~Li, X.~Tian, M.~Gong, Y.~Liu, T.~Liu, K.~Zhang, and D.~Tao, ``Deep domain generalization via conditional invariant adversarial networks,'' in \emph{Proceedings of the European Conference on Computer Vision (ECCV)}, 2018, pp. 624--639.

\bibitem{rahman2020correlation}
M.~M. Rahman, C.~Fookes, M.~Baktashmotlagh, and S.~Sridharan, ``Correlation-aware adversarial domain adaptation and generalization,'' \emph{Pattern Recognition}, vol. 100, p. 107124, 2020.

\bibitem{tobin2017domain}
J.~Tobin, R.~Fong, A.~Ray, J.~Schneider, W.~Zaremba, and P.~Abbeel, ``Domain randomization for transferring deep neural networks from simulation to the real world,'' in \emph{2017 IEEE/RSJ international conference on intelligent robots and systems (IROS)}.\hskip 1em plus 0.5em minus 0.4em\relax IEEE, 2017, pp. 23--30.

\bibitem{shankar2018generalizing}
S.~Shankar, V.~Piratla, S.~Chakrabarti, S.~Chaudhuri, P.~Jyothi, and S.~Sarawagi, ``Generalizing across domains via cross-gradient training,'' \emph{arXiv preprint arXiv:1804.10745}, 2018.

\bibitem{zhou2020deep}
K.~Zhou, Y.~Yang, T.~Hospedales, and T.~Xiang, ``Deep domain-adversarial image generation for domain generalisation,'' in \emph{Proceedings of the AAAI Conference on Artificial Intelligence}, vol.~34, no.~07, 2020, pp. 13\,025--13\,032.

\bibitem{wang2020heterogeneous}
Y.~Wang, H.~Li, and A.~C. Kot, ``Heterogeneous domain generalization via domain mixup,'' in \emph{ICASSP 2020-2020 IEEE International Conference on Acoustics, Speech and Signal Processing (ICASSP)}.\hskip 1em plus 0.5em minus 0.4em\relax IEEE, 2020, pp. 3622--3626.

\bibitem{zhou2021mixstyle}
K.~Zhou, Y.~Yang, Y.~Qiao, and T.~Xiang, ``Domain generalization with mixstyle,'' \emph{ICLR}, 2021.

\bibitem{bagnell2005robust}
J.~A. Bagnell, ``Robust supervised learning,'' in \emph{AAAI}, 2005, pp. 714--719.

\bibitem{sinha2017certifying}
A.~Sinha, H.~Namkoong, R.~Volpi, and J.~Duchi, ``Certifying some distributional robustness with principled adversarial training,'' \emph{arXiv preprint arXiv:1710.10571}, 2017.

\bibitem{gao2018robust}
R.~Gao, L.~Xie, Y.~Xie, and H.~Xu, ``Robust hypothesis testing using wasserstein uncertainty sets.'' in \emph{NeurIPS}, 2018, pp. 7913--7923.

\bibitem{zhu2020distributionally}
S.~Zhu, L.~Xie, M.~Zhang, R.~Gao, and Y.~Xie, ``Distributionally robust $ k $-nearest neighbors for few-shot learning,'' \emph{arXiv e-prints}, pp. arXiv--2006, 2020.

\bibitem{rahimian2019distributionally}
H.~Rahimian and S.~Mehrotra, ``Distributionally robust optimization: A review,'' \emph{arXiv preprint arXiv:1908.05659}, 2019.

\bibitem{ben2013robust}
A.~Ben-Tal, D.~Den~Hertog, A.~De~Waegenaere, B.~Melenberg, and G.~Rennen, ``Robust solutions of optimization problems affected by uncertain probabilities,'' \emph{Management Science}, vol.~59, no.~2, pp. 341--357, 2013.

\bibitem{duchi2016statistics}
J.~Duchi, P.~Glynn, and H.~Namkoong, ``Statistics of robust optimization: A generalized empirical likelihood approach,'' \emph{arXiv preprint arXiv:1610.03425}, 2016.

\bibitem{namkoong2016stochastic}
H.~Namkoong and J.~C. Duchi, ``Stochastic gradient methods for distributionally robust optimization with f-divergences,'' \emph{Advances in neural information processing systems}, vol.~29, 2016.

\bibitem{duchi2021learning}
J.~C. Duchi and H.~Namkoong, ``Learning models with uniform performance via distributionally robust optimization,'' \emph{The Annals of Statistics}, vol.~49, no.~3, pp. 1378--1406, 2021.

\bibitem{blanchet2019robust}
J.~Blanchet, Y.~Kang, and K.~Murthy, ``Robust wasserstein profile inference and applications to machine learning,'' \emph{Journal of Applied Probability}, vol.~56, no.~3, pp. 830--857, 2019.

\bibitem{lee2018minimax}
J.~Lee and M.~Raginsky, ``Minimax statistical learning with wasserstein distances,'' \emph{Advances in Neural Information Processing Systems}, vol.~31, 2018.

\bibitem{mohajerin2018data}
P.~Mohajerin~Esfahani and D.~Kuhn, ``Data-driven distributionally robust optimization using the wasserstein metric: Performance guarantees and tractable reformulations,'' \emph{Mathematical Programming}, vol. 171, no.~1, pp. 115--166, 2018.

\bibitem{staib2017distributionally}
M.~Staib and S.~Jegelka, ``Distributionally robust deep learning as a generalization of adversarial training,'' in \emph{NIPS workshop on Machine Learning and Computer Security}, vol.~1, 2017.

\bibitem{zhang2013domain}
K.~Zhang, B.~Sch{\"o}lkopf, K.~Muandet, and Z.~Wang, ``Domain adaptation under target and conditional shift,'' in \emph{International Conference on Machine Learning}.\hskip 1em plus 0.5em minus 0.4em\relax PMLR, 2013, pp. 819--827.

\bibitem{kuhn2019wasserstein}
D.~Kuhn, P.~M. Esfahani, V.~A. Nguyen, and S.~Shafieezadeh-Abadeh, ``Wasserstein distributionally robust optimization: Theory and applications in machine learning,'' in \emph{Operations Research \& Management Science in the Age of Analytics}.\hskip 1em plus 0.5em minus 0.4em\relax INFORMS, 2019, pp. 130--166.

\bibitem{villani2009optimal}
C.~Villani, \emph{Optimal transport: old and new}.\hskip 1em plus 0.5em minus 0.4em\relax Springer, 2009, vol. 338.

\bibitem{peyre2019computational}
G.~Peyr{\'e}, M.~Cuturi \emph{et~al.}, ``Computational optimal transport: With applications to data science,'' \emph{Foundations and Trends{\textregistered} in Machine Learning}, vol.~11, no. 5-6, pp. 355--607, 2019.

\bibitem{huber1965robust}
P.~J. Huber, ``A robust version of the probability ratio test,'' \emph{Annals of Mathematical Statistics}, vol.~36, no.~6, pp. 1753--1758, 1965.

\bibitem{zhou2005training}
Z.-H. Zhou and X.-Y. Liu, ``Training cost-sensitive neural networks with methods addressing the class imbalance problem,'' \emph{IEEE Transactions on knowledge and data engineering}, vol.~18, no.~1, pp. 63--77, 2005.

\bibitem{scott2012calibrated}
C.~Scott, ``Calibrated asymmetric surrogate losses,'' \emph{Electronic Journal of Statistics}, vol.~6, pp. 958--992, 2012.

\bibitem{aurelio2019learning}
Y.~S. Aurelio, G.~M. de~Almeida, C.~L. de~Castro, and A.~P. Braga, ``Learning from imbalanced data sets with weighted cross-entropy function,'' \emph{Neural processing letters}, vol.~50, no.~2, pp. 1937--1949, 2019.

\bibitem{xu2020class}
Z.~Xu, C.~Dan, J.~Khim, and P.~Ravikumar, ``Class-weighted classification: Trade-offs and robust approaches,'' in \emph{International Conference on Machine Learning}.\hskip 1em plus 0.5em minus 0.4em\relax PMLR, 2020, pp. 10\,544--10\,554.

\bibitem{courty2016optimal}
N.~Courty, R.~Flamary, D.~Tuia, and A.~Rakotomamonjy, ``Optimal transport for domain adaptation,'' \emph{IEEE transactions on pattern analysis and machine intelligence}, vol.~39, no.~9, pp. 1853--1865, 2016.

\bibitem{rabin2011wasserstein}
J.~Rabin, G.~Peyr{\'e}, J.~Delon, and M.~Bernot, ``Wasserstein barycenter and its application to texture mixing,'' in \emph{International Conference on Scale Space and Variational Methods in Computer Vision}.\hskip 1em plus 0.5em minus 0.4em\relax Springer, 2011, pp. 435--446.

\bibitem{bolley2007quantitative}
F.~Bolley, A.~Guillin, and C.~Villani, ``Quantitative concentration inequalities for empirical measures on non-compact spaces,'' \emph{Probability Theory and Related Fields}, vol. 137, no. 3-4, pp. 541--593, 2007.

\bibitem{shen2018wasserstein}
J.~Shen, Y.~Qu, W.~Zhang, and Y.~Yu, ``Wasserstein distance guided representation learning for domain adaptation,'' in \emph{Thirty-second AAAI conference on artificial intelligence}, 2018.

\bibitem{fang2013unbiased}
C.~Fang, Y.~Xu, and D.~N. Rockmore, ``Unbiased metric learning: On the utilization of multiple datasets and web images for softening bias,'' in \emph{Proceedings of the IEEE International Conference on Computer Vision}, 2013, pp. 1657--1664.

\bibitem{li2017deeper}
D.~Li, Y.~Yang, Y.-Z. Song, and T.~M. Hospedales, ``Deeper, broader and artier domain generalization,'' in \emph{Proceedings of the IEEE international conference on computer vision}, 2017, pp. 5542--5550.

\bibitem{ghifary2015domain}
M.~Ghifary, W.~Bastiaan~Kleijn, M.~Zhang, and D.~Balduzzi, ``Domain generalization for object recognition with multi-task autoencoders,'' in \emph{Proceedings of the IEEE international conference on computer vision}, 2015, pp. 2551--2559.

\bibitem{torralba2011unbiased}
A.~Torralba and A.~A. Efros, ``Unbiased look at dataset bias,'' in \emph{CVPR 2011}.\hskip 1em plus 0.5em minus 0.4em\relax IEEE, 2011, pp. 1521--1528.

\bibitem{he2016deep}
K.~He, X.~Zhang, S.~Ren, and J.~Sun, ``Deep residual learning for image recognition,'' in \emph{Proceedings of the IEEE conference on computer vision and pattern recognition}, 2016, pp. 770--778.

\bibitem{donahue2014decaf}
J.~Donahue, Y.~Jia, O.~Vinyals, J.~Hoffman, N.~Zhang, E.~Tzeng, and T.~Darrell, ``Decaf: A deep convolutional activation feature for generic visual recognition,'' in \emph{International conference on machine learning}.\hskip 1em plus 0.5em minus 0.4em\relax PMLR, 2014, pp. 647--655.

\bibitem{dou2019domain}
Q.~Dou, D.~Coelho~de Castro, K.~Kamnitsas, and B.~Glocker, ``Domain generalization via model-agnostic learning of semantic features,'' \emph{Advances in Neural Information Processing Systems}, vol.~32, 2019.

\bibitem{krizhevsky2012imagenet}
A.~Krizhevsky, I.~Sutskever, and G.~E. Hinton, ``Imagenet classification with deep convolutional neural networks,'' \emph{Advances in neural information processing systems}, vol.~25, 2012.

\bibitem{diamond2016cvxpy}
S.~Diamond and S.~Boyd, ``Cvxpy: A python-embedded modeling language for convex optimization,'' \emph{The Journal of Machine Learning Research}, vol.~17, no.~1, pp. 2909--2913, 2016.

\bibitem{flamary2021pot}
R.~Flamary, N.~Courty, A.~Gramfort, M.~Z. Alaya, A.~Boisbunon, S.~Chambon, L.~Chapel, A.~Corenflos, K.~Fatras, N.~Fournier \emph{et~al.}, ``Pot: Python optimal transport,'' \emph{Journal of Machine Learning Research}, vol.~22, no.~78, pp. 1--8, 2021.

\end{thebibliography}
\end{document}